\documentclass[sigconf, nonacm]{acmart}
\usepackage[utf8]{inputenc} 
\usepackage[T1]{fontenc}    
\usepackage{hyperref}       
\usepackage{url}            
\usepackage{booktabs}       
\usepackage{amsfonts}       
\usepackage{nicefrac}       
\usepackage{microtype}      
\usepackage{xcolor}         
\usepackage{stfloats}

\usepackage{graphicx}
\usepackage{subcaption}
\usepackage{booktabs} 
\usepackage{amsmath}
\usepackage{amsthm}

\usepackage{amssymb}
\usepackage{mathtools}
\usepackage{natbib}
\usepackage{thmtools}
\usepackage{thm-restate}
\usepackage{makecell}
\usepackage{booktabs}
\usepackage{multirow}
\usepackage{graphicx}
\usepackage{caption}
\usepackage{subcaption}
\usepackage{amsfonts}
\usepackage{algorithm}
\usepackage{algorithmic}
\usepackage{amssymb}
\usepackage{pifont}
\usepackage{xcolor}
\usepackage{makecell}
\usepackage{wrapfig}        
\definecolor{deepblue}{rgb}{0,0,0.5}
\definecolor{officeblue}{RGB}{0,102,204}
\definecolor{deepred}{rgb}{0.6,0,0}
\definecolor{deepgreen}{rgb}{0,0.5,0}
\definecolor{mybrickred}{RGB}{182,50,28}
\definecolor{fillcolor}{RGB}{216,217,252}

\renewcommand{\algorithmiccomment}[1]{\bgroup\hfill $\triangleright$ ~#1\egroup}

\definecolor{color_m}{RGB}{72,117,170}
\definecolor{color_f}{RGB}{201,89,72}
\definecolor{color_c}{RGB}{230,230,230}
\definecolor{color_e}{RGB}{100,155,74}
\definecolor{ccon}{HTML}{fee9d4}
\definecolor{cood}{HTML}{d8f0d3}
\definecolor{cid}{HTML}{dae8f5}
\definecolor{gg}{HTML}{e2f0cb}

\def\1{\bm{1}}

\DeclareMathAlphabet{\mathsfit}{\encodingdefault}{\sfdefault}{m}{sl}
\SetMathAlphabet{\mathsfit}{bold}{\encodingdefault}{\sfdefault}{bx}{n}

\def\Expectation{{\mathbb{E}}}

\newcommand{\Ls}{\mathcal{L}}

\theoremstyle{plain}

\theoremstyle{definition}

\theoremstyle{remark}

\usepackage{stfloats}

\newcommand{\INPUT}{\item[\textbf{Input:}]}
\newcommand{\OUTPUT}{\item[\textbf{Output:}]}

\newcommand{\bmu}{\text{\boldmath{$\mu$}}}

\newcommand{\bx}{\boldsymbol{x}}

\newcommand{\bs}{\boldsymbol{s}}

\newcommand{\btheta}{\boldsymbol{\theta}}
\newcommand{\bc}{\boldsymbol{c}}

\newcommand{\bI}{\boldsymbol{I}}

\newcommand{\diff}{\mathrm{d}}
\newcommand{\bwt}{\boldsymbol{w}_t}

\newcommand{\cP}{\mathcal{P}}

\AtBeginDocument{%
  }

\setcopyright{acmlicensed}
\copyrightyear{2026}
\acmYear{2026}
\acmDOI{XXXXXXX.XXXXXXX}
\acmConference[MM '26]{The 34th ACM International Conference on Multimedia}{November 10--14, 2026}{Rio de Janeiro, Brazil}
\acmISBN{978-1-4503-XXXX-X/2026/11}

\acmSubmissionID{}

\begin{document}

\title{OFA-Diffusion Compression: Compressing Diffusion Model in One-Shot Manner}

\author{Haoyang Jiang}
\authornote{Equal contribution.} 
\affiliation{%
  \institution{Renmin University of China}
  \city{Beijing}
  \country{China}
}
\email{jianghaoyang233@ruc.edu.cn}

\author{Zekun Wang}
\authornotemark[1] 
\affiliation{%
  \institution{Alibaba Inc.}
  \city{Hangzhou}
  \country{China}
}
\email{zkwang@ir.hit.edu.cn}

\author{Mingyang Yi}
\authornote{Corresponding author.} 
\affiliation{%
  \institution{Renmin University of China}
  \city{Beijing}
  \country{China}
}
\email{yimingyang@ruc.edu.cn}

\author{Xiuyu Li}
\affiliation{%
  \institution{Renmin University of China}
  \city{Beijing}
  \country{China}
}
\email{2022201508@ruc.edu.cn} 

\author{Lanqing Hu}
\affiliation{%
  \institution{Independent researcher}
  \country{} 
}
\email{hulanqing1@outlook.com}

\author{Junxian Cai}
\affiliation{%
  \institution{Tencent Inc.}
  \city{Shenzhen}
  \country{China}
}
\email{jasoncjxcai@tencent.com}

\author{Qingbin Liu}
\affiliation{%
  \institution{Tencent Inc.}
  \city{Shenzhen}
  \country{China}
}
\email{qingbinliu@tencent.com}

\author{Xi Chen}
\affiliation{%
  \institution{Tencent Inc.}
  \city{Shenzhen}
  \country{China}
}
\email{jasonxchen@tencent.com}

\author{Ju Fan}
\affiliation{%
  \institution{Renmin University of China}
  \city{Beijing}
  \country{China}
}
\email{fanj@ruc.edu.cn}

\vspace{5mm}


\begin{abstract}
The Diffusion Probabilistic Model (DPM) achieves remarkable performance in image generation, while its increasing parameter size and computational overhead hinder its deployment in practical applications. 
To improve this, the existing literature focuses on obtaining a smaller model with a fixed architecture through model compression. 
However, in practice, DPMs usually need to be deployed on various devices with different resource constraints, which leads to multiple compression processes, incurring significant overhead for repeated training.
To obviate this, we propose a once-for-all (OFA) compression framework for DPMs that yields different subnetworks with various computations in a one-shot training manner.
The existing OFA framework typically involves massive subnetworks with different parameter sizes, while such a huge candidate space slows the optimization. 
Thus, we propose to restrict the candidate subnetworks with a certain set of parameter sizes, where each size corresponds to a specific subnetwork.
Specifically, to construct each subnetwork with a given size, we gradually allocate the maintained channels by their importance.
Furthermore, we propose a reweighting strategy to balance the optimization process of different subnetworks.
Experimental results show that our approach can produce compressed DPMs for various sizes with significantly lower training overhead while achieving satisfactory performance. The code is available at \url{https://github.com/atrijhy/OFA-Diffusion_Compression}.
\end{abstract}





\maketitle

\section{Introduction}\label{sec:introduction}
Diffusion Probabilistic Models (DPMs)~\cite{sohl2015deep,ho2020denoising,song2021score} have attracted great attention in image synthesis due to their stable training paradigm and heightened controllability~\cite{dalle2,sd}.
Despite the superior performance of DPM, its inference process usually requires excessive computational overhead, due to the necessity of computing through complex architectures with numerous parameters (typical U-Nets~\citep{u-net}, DiT~\citep{dit}, and U-ViT \citep{uvit}) over multiple denoising steps. 
The massive inference cost hinders the deployment of DPMs on devices under specific efficiency constraints~\cite{chen2023speed}. 
\par
To mitigate this, the existing literature proposes to reduce the denoising steps by properly designed numerical samplers~\cite{ddim,dpm-solver,progressive-distillation,guided-distillation,consistency_model,unipc,xue2023sa,yi2024towards,wang2025improved,yi2023generalization}. 
Although these methods significantly reduce denoising steps, the volume of model parameters remains an efficiency bottleneck for each denoising step. 
Specifically, on edge devices, even a minimal number of evaluation steps can be slow or impractical. 
Therefore, compressing the architecture of DPMs, e.g., by pruning~\cite{snapfusion,diff-pruning}, knowledge distillation~\cite{bk-sdm}, or quantization~\cite{q-diffusion,ptqd} has become more crucial.
Nevertheless, the aforementioned compression methods compress the original DPM to a fixed size, while the resource constraints (e.g., parameter size, power, latency) vary widely in real-world deployment scenarios.
Thus, deploying a single fixed model across all types of devices is not feasible. 
Moreover, compressing specialized DPMs for each device requires multiple compression and retraining processes, which is computationally expensive and uneconomical.  
Thus, to amortize the expensive multiple compression overhead, it is desirable to compress DPMs into different sizes in a one-shot manner. 

To this end, we propose a \textit{once-for-all (OFA)} Diffusion Compression framework: jointly compresses different subnetworks in a one-shot training manner and adapts the computation flexibly by selecting the channel widths\footnote{Refers to the number of channels in layers.}.
Thus, the OFA compression not only significantly enriches the number of architectural configurations for deployment (more than dozens in our work), but also reduces the training overhead of subnetworks. 
Notably, OFA training has been proposed for classification networks or GAN networks. In existing methods, for each update step, the update direction can be obtained by enumerating all subnetworks~\cite{yu2019universally,dynabert,slimgan} or sample a random architectural configuration~\cite{layer-drop,anycost-gan}. 
However, neither is appropriate in our setting:
The former method is computationally impractical when scaling to our intended scales of subnetworks (more than dozens). 
The latter method has a huge candidate space, while we focus only on subnetworks that are at the Pareto frontier of performance and resource budgets in real-life deployment~\cite{stitch_network}, thus massive useless candidates are ignored.
Moreover, the huge space hinders the convergence of OFA training and requires an additional post-training search process.
Thus, to improve OFA training for DPMs, we propose to specify the subnetwork for each  size.

However, unifying the training process of different subnetworks with shared parameters is non-trivial, as the optimization of these subnetworks can negatively interfere with each other, leading to a slow convergence. Thus, the core problem becomes how to wisely construct the candidate subnetworks.
To mitigate negative interference and meanwhile fully utilize the information adopted from pre-trained DPM, we propose to keep more important channels to be shared across more subnetworks. 
Specifically, we first calculate the channel sensitivity criterion~\cite{taylor_importance} as its importance score. 
Then, for each target parameter size in training, we propose to prune channels in the same layer by ascending order of their importance scores, until the desired budget is attained. 
Taking into account the different importance of layers in DPM~\cite{bk-sdm}, we also introduce a strategy to maintain more parameters in more important layers.
After obtaining the desired candidate subnetworks, we simultaneously train them in a one-shot manner.
Moreover, considering the subnetworks of different sizes have different convergence rates during training, we propose a reweighting training strategy to balance the convergence of their optimization.  
Experimental results on various datasets show that our OFA diffusion compression framework only needs to be trained once while producing subnetworks with similar or even better performance compared to separately compressed models across various parameter sizes.


\begin{figure*}[t!]
\vspace{-0.1in}
    \begin{center}
    \includegraphics[width=0.95\linewidth]{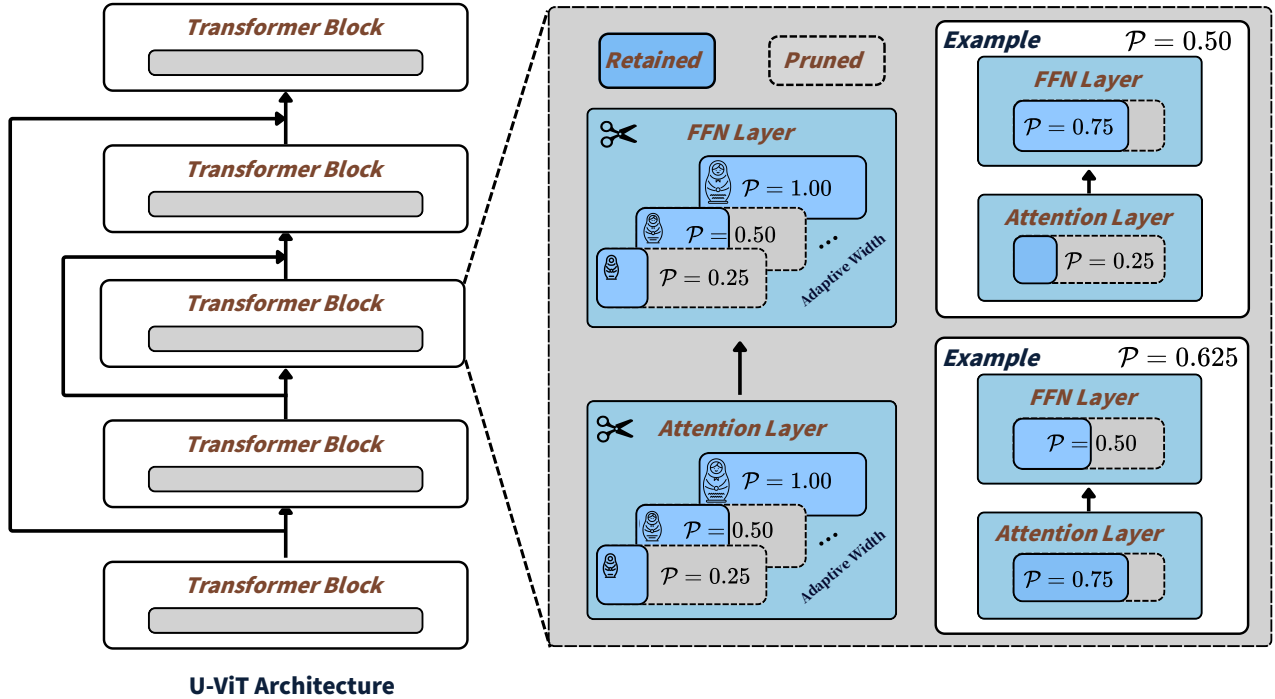}
    \caption{
    Overview of our OFA-Diffusion Compression framework.
    On the left, we present the structure of U-ViT architecture.
    On the right, we show the details of layers in blocks (FFN and Self-Attention) in our compression framework. We omit GroupNorm and time embedding for simplicity. Given a retention rate $\cP_{i}$, only the top-$\cP_{i}$ important parameters will be preserved in each block, then these budgets will be unevenly allocated to each layer in a greedy manner. By doing so, for two retention rates $\cP_{i} < \cP_{j}$, in each layer, the corresponded module of $\cP_{i}$ is a subset of the module corresponded to $\cP_{j}$.  
    }
    \label{figs:overview}
    \end{center}
\end{figure*}

\section{Related Work}

\paragraph{Model Compression of DPMs.}
Model compression aims to reduce the redundant parameters of large models and save the practical computations \citep{deep_compression}, including pruning~\citep{OBD, OBS, DBLP:journals/corr/HanPTD15}, knowledge distillation~\citep{hinton_kd,fitnets}, or quantization~\citep{xnor-net,hou2017loss-aware} techniques.
Considering the large computational overhead of DPM \citep{ho2020denoising,song2021score}, it is natural to apply these techniques to it. 
\citep{bk-sdm} drops redundant layer and utilizes knowledge distillation to mimic the original DPM.
\citep{diff-pruning} prunes unimportant channels based on training loss.
\citep{snapfusion} performs an evolutionary search to identify redundant blocks and then prunes the model until it satisfies the desired budget.
Leveraging quantization technique to accelerate DPMs has also been explored~\citep{ptq-diffusion, q-diffusion, ptqd, efficientdm}.
Although these works make great progress in compressing DPMs, the resulting single fixed model is not suitable for all deployment scenarios, as we have mentioned in Section \ref{sec:introduction}. 
Thus, we focus on compressing a pre-trained DPM into a range of smaller models with varied budgets (parameter sizes or latency) in a one-shot manner, which is critical in practice but remains unexplored. 


\paragraph{Once-for-All (OFA) Training}
To alleviate the excessive overhead of retraining from scratch for different models, the \textit{once-for-all} framework aims to train a single network that supports diverse architectural settings and selects only a part of the OFA network in inference based on device requirements~\citep{once-for-all}.
Slimmable network~\citep{slimmable_neural_networks, yu2019universally} is a typical OFA model that adjusts the network width configuration, permitting instant and adaptive performance-efficiency trade-offs at run-time.
\citep{anycost-gan, slimgan} extend this method to GANs for generative tasks.
On the other hand, \citep{dynabert} compresses BERT~\citep{bert} into smaller models with adaptive width and depth at once, and finally deploys them in various scenarios.
In this work, we generalize such OFA compression technique to DPM. Notably, \citep{oms-dpm}  independently train several DPMs of different sizes from scratch, applying proper structures during different steps of sampling process. The idea is similar to OFA compression, while their method is not scalable with the number of architectures. 
This is because the computational cost of training their DPMs are linearly increased with the number of candidate models. However, the subnetworks in our OFA compression framework are adopted from the same model, thus obviate the limitations in computational cost \citep{oms-dpm}.       
\section{Preliminaries}
\paragraph{Diffusion Probabilistic Models (DPMs).}
DPMs~\citep{sohl2015deep, ho2020denoising} are generative models that gradually introduce Gaussian noise to real data through a forward diffusion process and generate samples via a backward denoising process.
Following \citep{song2021score}, we introduce DPMs in a continuous-time manner. 
Generally, in the forward process, the noisy data distribution $p_{t}(\bx)$ is diffused from the real data $\bx_{0} \sim p_{\text{data}}(\bx)$, through a stochastic differential equation (SDE) \citep{oksendal2013stochastic}:
\begin{equation} 
\label{eq: forward process sde}
\diff \bx_t = \bmu(x_{t},t) \diff t + \sigma(t)  \diff \bwt,
\end{equation}
where $\bmu$ and $\sigma$ denote the drift and diffusion coefficients respectively, $\bwt$ is the standard Wiener process. For a large $T$, we can get $\bx_{T}\approx \mathcal{N}(0, \bI)$. Then, by running the following reverse time probability flow ordinary differential equation 
\begin{equation} 
\label{eq: reverse ODE}
\diff \bx_{t} = \left[\bmu(x_{t},t) - \frac{1}{2}\sigma(t)^{2} \nabla \log p_{t}(\bx_t)\right] \diff t,
\end{equation}
started from $\bx_{T}\approx \mathcal{N}(0, \bI)$ ($\bx_{T}$ is standard Gaussian in practice), we can sample $\bx_{0}\sim p_{\rm data}(\bx)$.

In this paper, we use the formulation of EDM~\citep{edm}, such that $\bmu(x_{t},t)=\mathbf{0}$ and $\sigma(t) = \sqrt{2t}$ with $\sigma(t) \in \left[0.002, 80\right]$.
A neural network $\bs_{\btheta}(\bx_t, t)$, is optimized to approximate the score function by minimizing the score matching loss~\citep{score_matching_loss}:
\begin{equation} 
\label{eq: score matching loss}
\Ls_{\btheta} \!\!=\!\! \Expectation_t \Bigl\{ \lambda(t) \Expectation_{\bx_{t}}  \bigl[\bigl\| \bs_{\btheta}(\bx_t, t) - \nabla_{\bx_t}\log p_{0t}(\bx_t|\bx_0) \bigl\|_2^2 \bigl] \Bigl\},
\end{equation}
where $\lambda(t)$ controls the importance weight of different time steps.
During sampling, the output of neural network $\bs_{\btheta}(\bx_t, t)$ replaces $\nabla \log p_{t}(\bx_t)$ in \eqref{eq: reverse ODE}.
The obtained samples $\hat{\bx}_0$ can be viewed as an approximation to data from $p_{\text{data}}(\bx)$. 


\paragraph{U-Net Architecture}
Generally, the U-Net~\citep{u-net} is the primary architecture (i.e., $\bs_{\btheta}$ in \eqref{eq: score matching loss}) in DPMs~\citep{ho2020denoising, dhariwal2021adm,sd}. 
The U-Net consists of $n$ pairs of downsampling-upsampling blocks $\{D_{i},U_{i}\}_{i=1}^{n}$, and another middle block $M$, where $i$ are distinguished by the resolution of the block.
At the end of $D_{i}$ (resp. $U_{i}$), a downsample (resp. upsample) convolution is inserted to reduce (resp. recover) the resolution of feature maps.
For each pair of blocks, there exist skip connections from $D_i$ to $U_{i}$ to transfer low-level information (see Appendix~A for a detailed illustration of the U-Net architecture).
In each layer of block ($D_{i}$, $U_{i}$ or $M$), there exist several stacked $3\times3$ convolution modules ($Conv_1$ and $Conv_2$), followed by group normalization \citep{group_norm}. These components consist of a ResNet layer \citep{resnet}. Besides that, some of these layers are followed by a spatial self-attention module~\citep{attention,pixelcnn++}.


\paragraph{Vision Transformer}
As an alternative to U-Net, Vision Transformer (ViT) based architectures~\citep{dit,uvit} replace the convolution layers with a pure transformer backbone as the score network $\bs_{\btheta}$. 
Concretely, for ViT-based architectures, the input image with resolution $H \times W$ is firstly patchified into a sequence of $N = \frac{HW}{p^2}$ tokens with patch size $p$. These tokens are then processed by a stack of $L$ transformer blocks $\{T_l\}_{l=1}^L$. Each block $T_l$ consists of a multi-head self-attention (MSA) module~\citep{attention} and a position-wise feed-forward network (FFN).
Unlike traditional convolution neural networks, transformer backbones apply global self-attention across all tokens at every layer to capture long-range dependencies.
In this work, we adopt U-ViT~\citep{uvit} as the representative ViT-based backbone for DPMs. 
As shown in the left part of Figure~\ref{figs:overview}, U-ViT further introduces long skip connections between shallow blocks $T_{l}$ and deep blocks $T_{L-l+1}$, similar in spirit to the $\{D_i, U_i\}$ skip connections in U-Net, to better capture low-level representations.



\section{Constructing Subnetworks Set}
\label{sec:construct_subnetwork}
In this section, we present details of the construction of subnetworks in our OFA-Diffusion Compression framework. 
The overview of the framework is shown in Figure~\ref{figs:overview}.
\par

In previous OFA work \citep{slimmable_neural_networks, once-for-all,layer-drop}, the subnetworks are randomly constructed at each update step. 
However, they produce massive subnetworks whose update direction can interfere with each other, slowing down the convergence of training and causing significant performance degradation on DPMs (see Figure~\ref{fig:ofa_comparison}) compared to subnetworks trained independently. 
\par
To accelerate the convergence of OFA training, we propose to reduce the number of candidate subnetworks (e.g., from thousands to dozens). It is worth noting that dozens of subnetworks in different sizes are enough to meet the practical deployment, since the deployment dedicate to find the subnetworks on the Pareto frontier of performance and resource budgets~\citep{stitch_network}, so that many useless subnetworks are ignored.    
Concretely, we specify a list of parameter retention rates (consisting of dozens of candidates): $\{\cP_{i}\}_{i=1}^{N}$, with $\cP_{1} < \cdots < \cP_{N}$, and each $\cP_{i}$ corresponds to a specific constructed subnetwork. 
Next, we describe the details of constructing these subnetworks.  


\subsection{Estimation of Channel Importance}
\label{sec:channel_importance}

In practice, compressing the DPMs either removes entire layers ~\citep{bk-sdm, snapfusion} or prune channels of features to reduce the width of model~\citep{diff-pruning}. However, removing coarse-grained units such as layers, usually leads to a significant performance drop~\citep{cofi}. 
Therefore, we focus on reducing the width of layers by pruning channels. Concretely, for each removed channel of features, the weights of compressed modules (convolutions and self-attention as in Figure~\ref{figs:overview}) connected to this channel will be eliminated following previous work~\citep{diff-pruning}.
To obtain subnetwork with retention rate $\cP_{i}$, for all blocks, we set their retention rates uniformly as $\cP_{i}$ \footnote{A model with retention rate $\cP_{i}$ can have different retention rates across blocks. However, we find that varying retention rates over blocks does not bring extra benefits. Thus, retention rates are invariant across  blocks.}. By doing this, we can directly control the actual computation overhead (e.g., MACs) with the retention rates of subnetworks. 

Then, we describe how to construct subnetworks with different channel widths.
Since subnetworks are trained jointly by sharing weights, it is intuitive to suggest that more important channels should be shared by more subnetworks to reduce interference between them.
To do this, we first estimate the importance of channels. 
As channel contributions to model performance have been observed to vary in the same layer~\cite{he2017channelpruning}, we use the sensitivity criterion~\citep{taylor_importance, PLATON} to measure the importance of channels, as in \citep{diff-pruning}. The criterion is obtained by the variation of training loss \eqref{eq: score matching loss} when the target channel is removed. 
Intuitively, the channel is considered important when this removal elicits a significant influence on the loss. 
Specifically, for the weight of the $i$-th channel $\bc_i$, its importance $I_{i}^c$ is approximated by the first-order Taylor expansion of $\Ls$ on $\bc_i=\mathbf{0}$:
\begin{equation}
\begin{aligned}
    I_{i}^C = \left|\Ls - \Ls_{\bc_i=\mathbf{0}}\right|
    = \left| (\bc_{i}-\mathbf{0})^\top \nabla \Ls + R_{\bc_{i}=\mathbf{0}}) \right|
    &\approx \left| \bc_{i}^\top \nabla\Ls \right|,
    \label{eq: compute importance}
\end{aligned}
\end{equation}
where $R_{\bc_{i}=\mathbf{0}}$ denotes the high-order remainder. 
In practice, we approximate $I^C$ in \eqref{eq: compute importance} with its empirical mean.

\subsection{Greedy Strategy for Channel Allocation among Layers}\label{sec:Greedy Allocation Strategy}
In the above section, we describe how to estimate the channel importance. 
In practice, the subnetwork architecture is obtained after specifying the number of channels maintained in each layer. 
Since the retention rate $\cP_{i}$ is invariant between blocks as mentioned before, our goal in the sequel is to specify the channel allocation in each layer of blocks.
\par
In the existing literature \citep{slimmable_neural_networks, dynabert, slimgan} the retention rates are also invariant among layers. However, the importance of layers in the same block varies significantly in DPMs~\citep{bk-sdm} (Figure~8 in Appendix), which may cause the invariant strategy to be suboptimal for channel allocation. 
Therefore, we propose a greedy strategy that flexibly allocates the number of channels among layers based on layer importance scores.
\par
\begin{figure}[t!]
    \centering
    \includegraphics[width=\columnwidth]{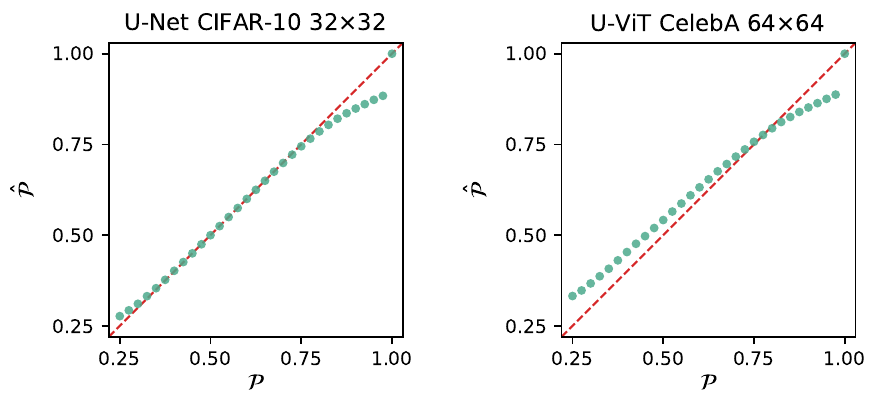}
    \caption{The practical retention rates of parameters $\hat{\cP}_{i}$ over 
    the ideal ones $\cP_{i}$, under the structures of U-Net and U-ViT used in Section \ref{sec:exp}.}
    \label{fig:pvsp}
\end{figure}

To do this, for the layer $l$ of a block with $K$ layers in total, we calculate its importance score $I^{L}_l$, by aggregating the importance of all channels in it: $I^{L}_l = \sum_{\bc_i \in l} I_{\bc_i}^{c}$. 
We observe that the layer importance $I^{L}$ generally match the performance drop of eliminating layers (see Appendix~D).
With these importance scores of the layers, we proceed to allocate the channels maintained in each layer. 
As the retention rate $\cP_{i}$ is invariant across blocks, the number of channels maintained in each block is $K\cP_{i}|l|$, where $|l|$ is the original number of channels in each layer of block. 
Based on $I_{l}^L$, we propose to maintain
\begin{equation}\label{eq:allocation}
    C_{l} = \max\left\{\min\left\{\left\lceil\frac{I_l^L}{\sum_{l} I_l^L} K\cP_{i}|l|\right\rceil, |l|\right\}, \lceil\cP_{0}|l|\rceil\right\}
\end{equation}
channels in layer $l$, where the $\min$ and $\max$ operations are respectively to guarantee the upper and lower bounds (the lower bound guarantees no layers are entirely removed) of maintained channels in compressed layers. After that, the retention rate for each module in this layer is set as $|C_{l}|/|l|$.\footnote{Theoretically, the retention rates of modules in a layer can be varied. However, we find that keeping them invariant is more computational efficient in practice.} Notably, due to \eqref{eq:allocation}, the resulted subnetwork corresponds to $\cP_{i}$ has practical compression rate $\hat{\cP}_{i}$ that may differ from $\cP_{i}$. However, in practice, we find that $\{\hat{\cP}_{i}\}_{i=1}^{N}$ are extremely close to $\{\cP_{i}\}_{i=1}^{N}$, as shown in Figure~\ref{fig:pvsp}.

The overall process of subnetwork construction is shown in Algorithm~\ref{alg: construction}. It is worth noting that in our construction, the smaller subnetworks are part of the larger ones, which naturally mitigates the interference between them.

\section{Model Training}\label{sec:training}
With constructed subnetworks as in Section \ref{sec:construct_subnetwork}, we are ready to train them in one-shot manner. 
The target of our OFA compression training is to optimize all supported subnetworks, rendering their performance commensurate with that achieved by separate training.
We propose training them by the score matching loss \eqref{eq: score matching loss} as the original DPM. 
Ideally, in each update step, we randomly sample a retention rate $\cP_{i}$, and update the parameters of the corresponding subnetwork under $\cP_{i}$ \footnote{Please note that the subnetworks have shared weights and subnetworks are constructed by parameters masks during training, so that the memory overhead will not exceed the standard training process.}. 
Next, we denote the corresponding parameters of $\cP_{i}$ as $\btheta_{\cP_{i}}$. 

It is worth noting that the parameters of all subnetworks are initialized from a pre-trained DPM, thus intuitively, a larger $\cP_{i}$ leads to a faster convergence rate of its corresponding subnetwork $\btheta_{\cP_{i}}$\footnote{The convergence rate of subnetworks are in Appendix~C}.
Consequently, to improve training efficiency, we can balance the convergence progress of subnetwork optimization. 
We propose to minimize the following reweighting loss in OFA training:
\begin{equation}
    \min_{\btheta}\mathbb{E}_{\cP\in\{\cP_{i}\}}\left[\mathcal{L}_{\btheta_{\cP}}\right] = \sum_{i=1}^{N}w_{\cP_{i}}\mathcal{L}_{\btheta_{\cP_{i}}},
\end{equation}
where $\cP$ is a discrete distribution over $\{\cP_{i}\}_{i=1}^{N}$, and $w_{\cP_{i}}$ is the probability of $\cP$ takes value of $\cP_{i}$. 
As mentioned above, since a larger $\cP_{i}$ corresponds to a faster convergence rate, we propose to make $w_{\cP_{i}}$ decrease with $\cP_{i}$. Concretely, we make $w_{\cP_{i}}$ linearly descend with $i$, and make $w_{\cP_{1}} = mw_{\cP_{N}}$ \footnote{These $w_{\cP_{i}}$ are normalized with $\sum_{i} w_{\cP_{i}} = 1$.} for a constant $m$ (e.g., $m=3$).  
\par
The reweighting strategy is similarly appeared in the existing literature of OFA training. For example,  \citep{yu2019universally} proposed a \textit{sandwich} strategy whereas $w_{\cP_{1}} = w_{\cP_{N}} = 0.25$, and the other $w_{\cP_{i}}$ are the same constant. However, the strategy is proposed to be applied under training from scratch, and is suboptimal for OFA compression: assigning 25\% weights respectively on $\btheta_{\cP_{1}}$ and $\btheta_{\cP_{N}}$ leads to overfitted $\btheta_{\cP_{N}}$ and underfitted other subnetworks.
Section~\ref{sec:ablation study} explores the training strategy and verifies that our proposed strategy performs better.   
\begin{algorithm}[t!]
  \caption{Subnetwork Construction Process}\label{alg: construction}
  \begin{algorithmic}[1]
    \INPUT {
        $\btheta$: parameters of pre-trained model, $\left[\cP_i\right]_{i=1}^{N}$: retention rates list.
    }
    \STATE Compute channel importance $I_{i}^{C}$ in \eqref{eq: compute importance}. 
    \FOR{$\cP_i$ in $\left\{\cP_i\right\}$}{
        \FOR{block $\mathcal{B}$ ($D_{i}, U_{i}, M$ of Figure \ref{figs:overview}) in $\btheta$}{
            \STATE Compute layer importance $I^{l}$ for layer $l$ in $\mathcal{B}$: $I^{L}_l = \sum_{\bc_i \in l} I_{\bc_i}^{c}$
            \FOR{layer $l$ in $\mathcal{B}$}{
            \STATE Maintain channels with the top-$\max\left\{\min\left\{\frac{I_l}{\sum_{l} I_l} K\cP_{i}|l|, |l|\right\}, \cP_{0}|l|\right\}$ channels importance.  
            }\ENDFOR
            
        }\ENDFOR
        \STATE Obtain subnetwork $\btheta_{\cP_{i}}$ with retention rates $\cP_{i}$.
    }\ENDFOR
    \OUTPUT {
       Subnetworks $\left\{\btheta_{\cP_i}\right\}$.
    }
  \end{algorithmic}
\end{algorithm}
\section{Experiments}\label{sec:exp}

\subsection{Experimental Setup}
\begin{figure*}
    \centering
    \includegraphics[width=0.9\textwidth]{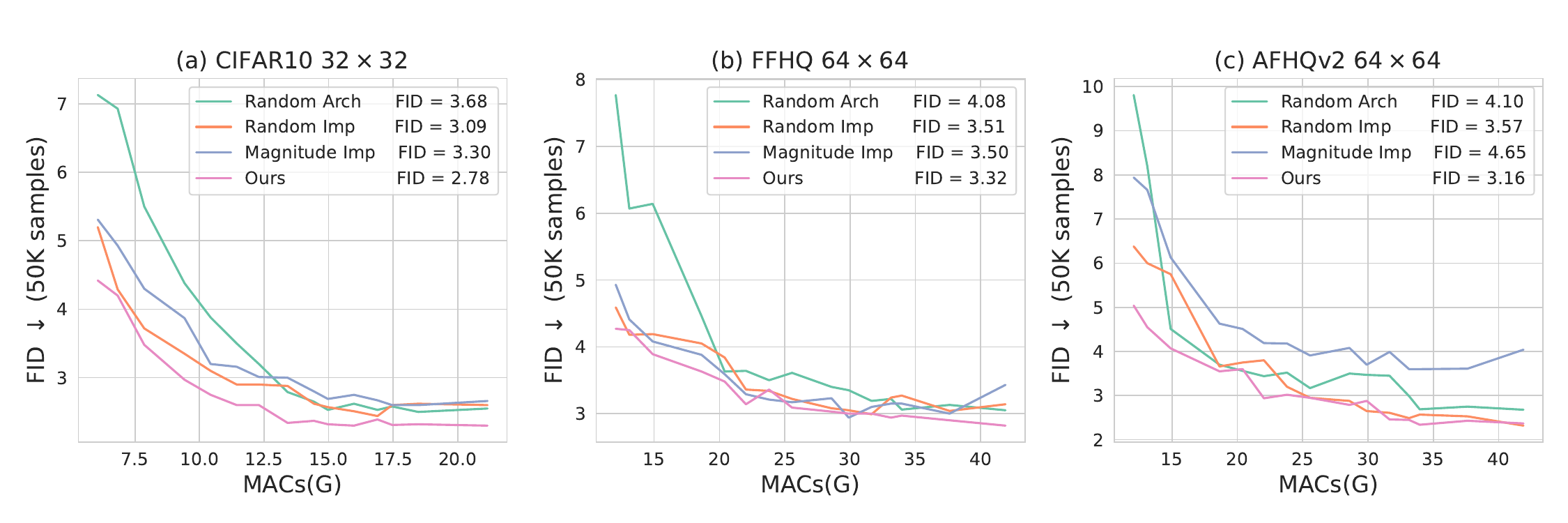}
    \caption{Comparison between our proposed subnetwork construction approach and other ablations. 
    For simplicity, ``Arch'' and ``Imp'' denote the ``Architecture'' and ``Importance'', respectively.
    All OFA networks are under the same training and evaluation settings. The results are obtained under U-Net, all the FID scores of 28 subnetworks are plotted. 
    We also report the averaged FID across all subnetworks for each method.
    We can observe that our method consistently achieves a better averaged FID on all datasets.
    }
    \vspace{-0.05in}
    \label{fig:ofa_comparison}
\end{figure*}
\paragraph{Models.} Our experiments are conducted on three representative diffusion models covering different architectures and generation tasks: Elucidating Diffusion Models (EDM)~\citep{edm} based on the standard U-Net architecture, U-ViT~\citep{uvit} utilizing the transformer architecture, and Stable Diffusion (SD)~\citep{sd} for the text-to-image generation task.

\paragraph{Datasets.}
To evaluate U-Net-based architectures, we conduct experiments on CIFAR10 $32\times32$~\citep{Krizhevsky09learningmultiple}, FFHQ $64\times64$~\citep{ffhq}, and AFHQv2 $64\times64$~\citep{afhqv2} for unconditional generation, alongside ImageNet $64\times64$~\cite{imagenet} for class-conditional tasks. For transformer-based architectures, we employ CIFAR10 $32\times32$ and CelebA $64\times64$~\citep{celeba}. Furthermore, text-to-image generation is evaluated on MS-COCO 2014~\citep{coco}, generating $512\times512$ images from validation captions.
\paragraph{Evaluation.}
We use Frechet Inception Distance (FID) score~\citep{fid} to evaluate the quality of generation.
Unless otherwise specified, 50K images are sampled to evaluate the FID score. 
We also present both the parameter retention rates $\cP$ and MACs (Multiply-Accumulate operations) of the corresponding subnetworks, where the MACs are measured under one number of function evaluation (NFE=1) with batch size of 1.

\paragraph{Implementation Details.}

During OFA training, we construct the search space at the layer level. Specially, different layers can independently choose from a predefined set of discrete retention rates (i.e., 4 candidates). Although this design induces a large combinatorial space of subnetworks, many configurations lead to the same overall parameter count. To calculate importance scores, we sample 1024 pairs $(\bx_0, t)$ from training datasets. Our models are mainly U-Net and ViT. 
\begin{enumerate}
    \item For U-Net, our approach is based on EDM~\citep{edm} and Stable-Diffusion v1.5 \citep{sd} (Text-to-Image generation). The parameters of our subnetworks are adopted from the  pre-trained backbones. After removing the  duplicated subnetworks in same size, we obtain 28 unique overall parameter retention rates $\{\cP_i\}_{i=1}^{28}$, covering the range from 0.25 to 1.0. Our method applied to U-Net refers to Figure~6 in Appendix. 
    \item For transformer-based architectures (ViT), we use U-ViT \citep{uvit}, which has 19 $\{
    \cP_{i}\}_{i=1}^{19}$ sizes of subnetworks after removing duplications, ranging from 0.25 to 1.0. The application to U-ViT refers to Figure \ref{figs:overview}.     
\end{enumerate}
More details of hyperparameters are shown in Appendix~B.


\begin{table*}[t!]
\caption{
FID scores ($\downarrow$) on CIFAR10 $32\times32$, FFHQ $64\times64$, AFHQv2 $64\times64$ and ImageNet $64\times64$.
$N$ is the number of subnetworks (28 in our experiments) while we show 7 subnetworks here for space limits. 
The training overhead of our OFA network remains constant as the number of compressed subnetworks increases.
``+Fine-tune" denotes the specific subnetworks are further fine-tuned after grabbing weights from our OFA network.
$\dag$ are the performance of pre-trained models. We mark the best results under each $\cP_i$ in \bf{bold}.
} 
\small\centering
\scalebox{0.95}{
\begin{tabular}{cl|lcccccccccccc|c}
\toprule
\multirow{3}{*}{Datasets} & \multirow{3}{*}{Method} & $\cP$ & $0.25\times$ & $0.375\times$ & $0.5\times$ & $0.625\times$ & $0.75\times$ & $0.875\times$ & $1.0\times$ & \multirow{2}{*}{Avg.} & \multirow{2}{*}{Training Overhead} \\ 
& & MACs & 6.06G & 9.42G & 11.44G & 13.41G & 15.98G & 18.47G & 21.15G \\
& & & FID & FID & FID & FID & FID  & FID & FID & FID & (K iterations)\\
\midrule
\multirow{4}{*}{CIFAR10 $32\times32$} & Separate Compression~\citep{diff-pruning}  & & 3.60 & 2.74 & 2.50 & 2.28 & 2.21 & 2.11 & 1.97$^{\dag}$ & 2.49 & $200N$\\
& Naive OFA~\citep{once-for-all} & & 5.31 & 3.87 & 3.16 & 3.00 & 2.75 & 2.60 & 2.66 & 3.34 & 200\\
 & Ours & & 4.42 & 2.97 & 2.60 & 2.34 & 2.30 & 2.32 & 2.30 & 2.75 & 200\\
 & \quad + Fine-tune 10K & & 3.45 & 2.64 & 2.36 & 2.28 & 2.15 & 2.05 & 1.95 & 2.41 & $200+10N$\\
 & \quad + Fine-tune 20K & & \bf{3.29} & \bf{2.52} & \bf{2.32} & \bf{2.14} & \bf{2.00} & \bf{1.96} & \bf{1.85} & \bf{2.29} & $200+20N$\\

\midrule
\midrule
& & MACs & 12.06G & 14.89G & 22.00G & 28.83G & 31.95G & 38.78G & 41.9G \\
\midrule
\multirow{4}{*}{FFHQ $64\times64$} & Separate Compression~\citep{diff-pruning} & & 4.27 & 3.60 & 3.26 & 3.12 & 2.83 & 2.74 & \bf{2.39}$^{\dag}$ & 3.17 & $200N$\\
& Naive OFA~\citep{once-for-all} & & 4.93 & 4.08 & 3.29 & 3.23 & 3.10 & 3.00 & 3.43 & 3.58 & 200\\
 & Ours & & 4.27 & 3.63 & 3.14 & 3.03 & 2.88 & 2.90 & 2.82 & 3.24 & 200\\
 & \quad + Fine-tune 10K & & 4.18 & 3.30 & 3.00 & \bf{2.84} & \bf{2.77} & 2.70 & 2.61 & 3.06 & $200+10N$\\
 & \quad + Fine-tune 20K & & \bf{4.06} & \bf{3.27} & \bf{2.98} & 2.92 & 2.79 & \bf{2.66} & 2.61 & \bf{3.04} & $200+20N$\\

\midrule
\midrule
& & MACs & 12.06G & 14.89G & 22.00G & 28.83G & 31.95G & 38.78G & 41.9G \\
\midrule
\multirow{4}{*}{AFHQv2 $64\times64$} & Separate Compression~\citep{diff-pruning} & & 5.66 & 4.07 & 3.15 & 3.12 & \bf{2.34} & 2.77 & \bf{1.96}$^{\dag}$ & 3.32 & $200N$\\
& Naive OFA~\citep{once-for-all} & & 7.94 & 6.13 & 4.19 & 3.60 & 3.99 & 3.61 & 4.04 & 4.78 & 200\\
 & Ours & & \bf{5.04} & 3.55 & 2.94 & 2.79 & 2.46 & 2.43 & 2.37 & 3.05 & 200 \\
 & \quad + Fine-tune 10K & & 5.23 & 3.52 & 2.88 & 2.72 & 2.48 & 2.40 & 2.29 & 3.07 & $200+10N$\\
 & \quad + Fine-tune 20K & & 5.30 & \bf{3.38} & \bf{2.82} & \bf{2.67} & 2.56 & \bf{2.31} & 2.02 & \bf{3.00} & $200+20N$\\
 \midrule
 \midrule
 
 & & MACs & 32.35G & 42.63G & 55.30G & 68.33G & 83.78G & 98.46G & 107G \\
 \midrule
\multirow{4}{*}{ImageNet $64\times64$} & Separate Compression~\citep{diff-pruning} & & 9.45 & 7.64 & 7.02 & 5.84 & 5.28 & 4.77 & \bf{2.57}$^{\dag}$ & 6.08 & $400N$\\
& Naive OFA~\citep{once-for-all} & & 14.04 & 10.25 & 7.81 & 7.26 & 6.85 & 6.29 & 6.00 & 8.36 & 400\\
 & Ours & & 11.35 & 9.13 & 7.10 & 6.60 & 6.12 & 5.95 & 5.75 & 7.43 & 400 \\
 & \quad + Fine-tune 20K & & 9.08 & 7.20 & 6.43 & 5.57 & 5.32 & 4.65 & 4.12 & 6.05 & $400+20N$\\
 & \quad + Fine-tune 40K & & \bf{8.65} & \bf{6.84} & \bf{5.08} & \bf{4.84} & \bf{4.97} & \bf{4.30} & 3.65 & \bf{5.47} & $400+40N$\\
\bottomrule
\end{tabular}
}
\label{tab:main}
\end{table*}


\subsection{Main Results}
Table~\ref{tab:main} shows the results of subnetworks derived from our OFA diffusion compression framework.
Due to space limits, we show the results of 7 subnetworks here\footnote{We visualize some of the performances of all subnetworks extracted from our OFA model in Figure~\ref{fig:ofa_comparison}.}. 
In Table \ref{tab:main}, we compare our method with a competitive baseline, \textit{Separate Compression}, which compresses subnetworks as in \citep{diff-pruning}, with specific parameter retention rates $\cP_{i}$ separately. In this regime, the compressed networks do not interfere with each other, while the training overhead increases linearly with the number of models. 
Besides, we also consider another baseline: the naive implementation of OFA framework as in ~\citep{once-for-all}.
\par
As shown in Table~\ref{tab:main}, our OFA network achieves comparable or even better performance (e.g., averaged FID with 3.05 vs. 3.32 on AFHQv2) with the separate compression baseline across various $\cP_{i}$, while enjoying a significant reduction in training overhead (e.g., $N \times 200$K iterations vs. 200K iterations). 
Moreover, our approach also significantly improves the naive implementation of OFA training~\citep{once-for-all} for DPMs.
In addition, with a slight further fine-tuning, subnetworks extracted from our OFA networks outperform the baselines on most retention rates and still have significantly fewer training computations. 
Notably, in a smaller retention rate regime ($\cP_{i} < 0.5$), our OFA approach beats separate compression by a large margin, suggesting that our method is more suitable in deployment scenarios with limited computational resources.
Furthermore, due to the weight-sharing property, our OFA network has less disk storage than the separate compression baseline (e.g., 213M vs. 3578M on CIFAR10).  
Interestingly, unlike other architectures, OFA compressed networks for Stable Diffusion already achieve highly competitive performance without fine-tuning. Moreover, further fine-tuning does not lead to any additional improvements, which we conjecture is primarily caused by overfitting~\cite{dreambooth,svdiff}.

In summary, our OFA diffusion compression framework is able to produce a wide range of compressed subnetworks with competitive performance in a computationally cheap one-shot manner, which helps us to better explore performance-efficiency trade-offs in practical deployment scenarios.  
The comparison of generated images by these subnetworks are in Appendix~F. 

\begin{figure}[t]
    \centering
    \includegraphics[width=\columnwidth]{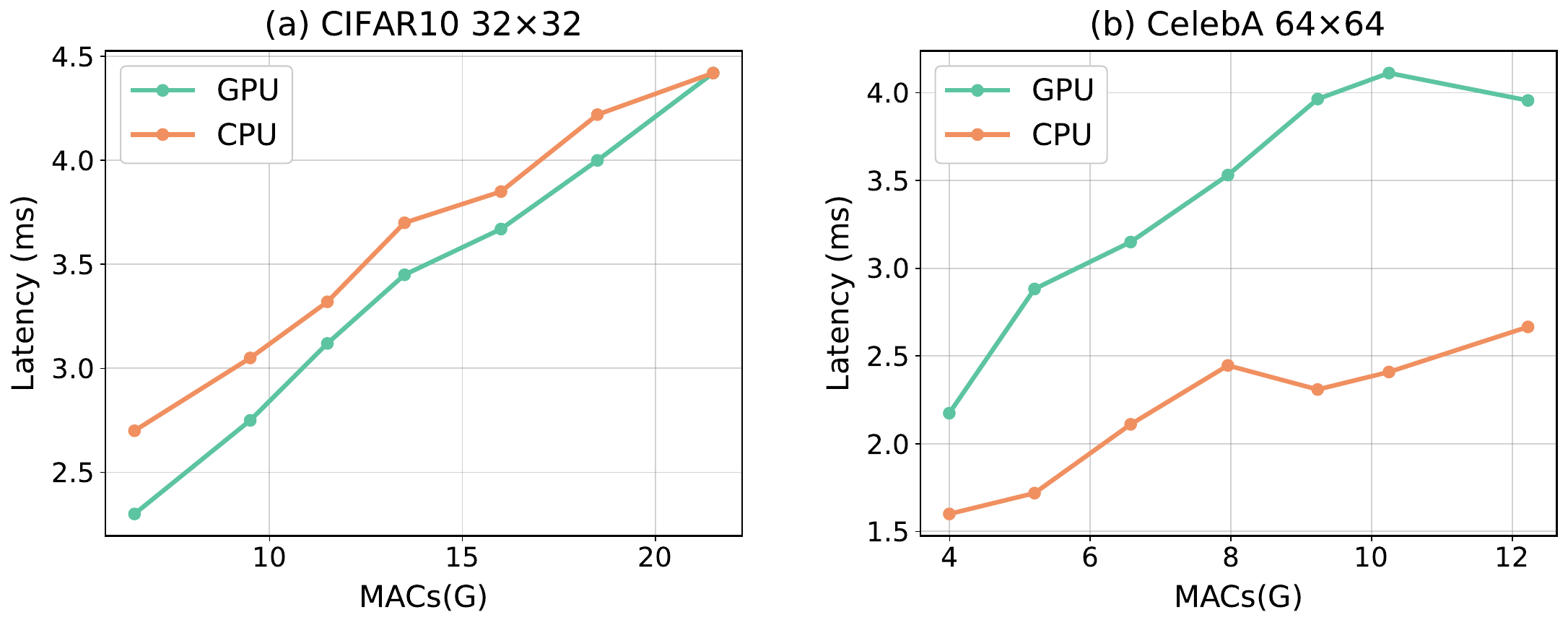}
    \caption{Averaged latency of model on GPU or CPU evaluated on
    (a) U-Net CIFAR-10 $32\times32$ with 35-step Heun's sampler and
    (b) U-ViT CelebA $64\times64$ with 50-step DPM-Solver.
    The GPU and CPU latency are evaluated with a batch size of 64 and 1, respectively.}
    \label{fig:latency}
    \vspace{-0.1in}
\end{figure}

\paragraph{Latency Reduction.}
We further measure the practical latency of subnetworks under various $\cP_{i}$ on both GPU and CPU devices.
The results are evaluated on a NVIDIA Tesla V100 GPU and an Intel Xeon Gold 6151 CPU, respectively.
As shown in Figure~\ref{fig:latency}, we observe that the subnetworks extracted from the OFA network indeed accelerate the original model. 
This consistent acceleration is observed on both the traditional U-Net (Figure~\ref{fig:latency}a) and the U-ViT architecture (Figure~\ref{fig:latency}b). 
We find that the retention rate $\cP_{i}$ is not exactly the inverse of the actual latency reduction ratio (e.g., $25\%$ vs. $2.4\times$), which also appears in previous channel pruning work~\citep{cofi, simple, diff-pruning}. This is because the computation of layer is conducted in parallel, so that reducing the channel width of layers does not linearly decrease the latency. 
In fact, compressing depth linearly reduce latency, but significantly sacrifices performance~\citep{cofi}. 
On the other hand, reducing the width by pruning channels as ours can save memory footprint in the layer-by-layer inference setting~\citep{slimgan}. Thus, considering all of these, we suggest pruning channels as ours in practice.

\subsection{Ablation Study}\label{sec:ablation study}
Next, we conduct a series of ablation studies to explore the effectiveness of key components in our OFA diffusion compression framework. 
For efficiency, we ablate on unconditional generation datasets.

\begin{table*}[t!]
\caption{
FID scores ($\downarrow$) on CIFAR10 $32\times32$ and CelebA $64\times64$ with a U-ViT backbone, and on MS-COCO2014 $512\times512$ with a Stable Diffusion v1.5 backbone.
}
\small\centering
\scalebox{0.9}{
\begin{tabular}{cl|lcccccccc|c}
\toprule
\multirow{3}{*}{Datasets} & \multirow{3}{*}{Method} & $\cP$ & $0.25\times$ & $0.375\times$ & $0.5\times$ & $0.625\times$ & $0.75\times$ & $0.875\times$ & $1.0\times$ & \multirow{2}{*}{Avg.} & \multirow{2}{*}{Training Overhead} \\
& & MACs & 3.09G & 4.28G & 5.71G & 7.12G & 8.46G & 9.64G & 11.39G & \\
& & & FID & FID & FID & FID & FID & FID & FID & FID & (K iterations) \\
\midrule
\multirow{5}{*}{CIFAR10 $32\times32$}
& Separate Compression~\citep{diff-pruning} & & 6.83 & \bf{5.58} & 5.28 & 5.39 & 5.55 & 5.63 & \bf{3.73}$^{\dag}$ & 5.43 & $50N$ \\
& Naive OFA~\citep{once-for-all}            & & 20.96 & 11.41 & 8.52 & 8.56 & 8.18 & 7.82 & 7.38& 10.40 & 50 \\
& Ours                                      & & 12.58 & 9.46 & 7.89 & 7.35 & 6.96 & 6.66 & 6.49 & 8.20 & 50 \\
& \quad + Fine-tune 10K                     & & 8.99 & 7.60 & 7.03 & 6.63 & 6.34 & 6.03 & 5.74 & 6.91 & $50+10N$ \\
& \quad + Fine-tune 20K                     & & \bf{5.91} & 6.24 & \bf{5.01} & \bf{4.71} & \bf{5.34} & \bf{5.46} & 3.87 & \bf{5.22} & $50+20N$ \\
\midrule
\midrule
& & MACs & 4.00G & 5.21G & 6.57G & 7.96G & 9.23G & 10.25G & 12.23G & \\
\midrule
\multirow{5}{*}{CelebA $64\times64$}
& Separate Compression~\citep{diff-pruning} & & \bf{3.67} & \bf{2.95} & 2.77 & 2.62 & 2.50 & 2.41 & 2.87$^{\dag}$ & 2.82 & $100N$ \\
& Naive OFA~\citep{once-for-all}            & & 10.93 & 4.96 & 3.97 & 3.35 & 3.19 & 2.96 & 2.91 & 4.61 & 100 \\
& Ours                                      & & 5.92 & 4.06 & 2.87 & 2.68 & 2.58 & 2.45 & 2.46 & 3.29 & 100 \\
& \quad + Fine-tune 10K                     & & 4.72 & 3.50 & 2.83 & 2.62 & 2.51 & 2.39 & 2.33 & 2.98 & $100+10N$ \\
& \quad + Fine-tune 20K                     & & 4.04 & 3.06 & \bf{2.73} & \bf{2.51} & \bf{2.42} & \bf{2.37} & \bf{2.32} & \bf{2.78} & $100+20N$ \\
\midrule
\midrule
& & MACs & 178.68G & 206.77G & 234.65G & 262.14G & 282.86G & 309.27G & 338.75G & \\
\midrule
\multirow{3}{*}{MS-COCO $512\times512$}
& Separate Compression~\citep{diff-pruning} & & 11.90 & 10.85 & 10.37 & 9.85 & 9.62 & 9.41 & 8.71$^{\dag}$ & 10.10 & $50N$ \\
& Naive OFA~\citep{once-for-all}            & & 11.64 & 9.47 & 8.97 & \bf{8.59} & \bf{8.42} & \bf{8.37} & 8.53 & 9.14 & 50 \\
& Ours                                      & & \bf{9.58} & \bf{9.26} & \bf{8.96} & 8.68 & 8.47 & 8.54 & \bf{8.44} & \bf{8.85} & 50 \\

\bottomrule
\end{tabular}
}
\label{tab:additonal}
\end{table*}


\paragraph{Effects of Subnetworks Construction}
As described in Section \ref{sec:construct_subnetwork}, for each parameter retention rate $\cP_{i}$, we specify a target subnetwork constructed based on importance scores of channels in the DPM.
We explore the effectiveness of the construction approach Algorithm \ref{alg: construction} we devised.
First, to investigate the impact of reducing the candidate space in OFA training, we compare our approach with a method adopted in the existing literature~\citep{once-for-all, layer-drop, anycost-gan}, \textit{Random Architecture}, which randomly constructs a subnetwork in each iteration.
For each $\cP_{i}$, unlike specifying one subnetwork as ours, the \textit{Random Architecture} has plenty of candidate architectures and requires a post-training search stage to find the optimal candidate\footnote{For a fair comparison, we also incorporate the search stage in implementing the \textit{Random Architecture} ablation, but note that the post-training search also introduce more computation overhead.}.
\par
Recall that in our approach, the importance scores are defined by the corresponding sensitivity metric \eqref{eq: compute importance}. 
We compare our approach with alternative strategies, where channel importance scores are randomly assigned (\textit{Random Importance}) or calculated according to the $L_{1}$ norm of channel weight (\textit{Magnitude Importance})~\citep{deep_compression}, which widely used in previous OFA work on classification task or GANs.
For a fair comparison, all variants employ the same training and evaluation settings.
Figure~\ref{fig:ofa_comparison} shows the performance of our OFA-Diffusion Compression with other ablations. As can be seen:
\begin{enumerate}
    \item \textit{Random Architecture} has the worst performance compared with the other methods, demonstrating that reducing the number of useless subnetworks can significantly improve the OFA training.
    \item Our approach consistently achieves better averaged FID scores over \textit{Random Importance} and \textit{Magnitude Importance} on all datasets, suggesting that our sensitivity-based importance score constructs better subnetworks.

\end{enumerate}

\paragraph{Effects of Reweighting Strategy}
\begin{table*}[t!]
\caption{
Ablation of training strategy in OFA-Diffusion Compression framework evaluated on U-Net. The best results for each $\cP_i$ are \bf{bold} marked.
} 
\small\centering
\scalebox{1.2}{
\begin{tabular}{cl|cccccccc}
\toprule
Datasets & Method & $0.25\times$ & $0.375\times$ & $0.5\times$ & $0.625\times$ & $0.75\times$ & $0.875\times$ & $1.0\times$ & Avg. \\
\midrule
 \multirow{3}{*}{CIFAR10 $32\times32$} & Uniform & 6.38 & 3.62 & 3.01 & 2.55 & 2.48 & 2.35 & 2.34 & 3.25 \\
 & Sandwich & 4.71 & 3.30 & 3.21 & 2.68 & 2.70 & \bf{2.29} & 2.60 & 3.07 \\
 & Ours & \bf{4.42} & \bf{2.97} & \bf{2.60} & \bf{2.34} & \bf{2.30} & 2.32 & \bf{2.30} & \bf{2.75} \\

\midrule
\midrule
 \multirow{3}{*}{FFHQ $64\times64$} & Uniform & 5.10 & 3.95 & 3.08 & 3.15 & 2.78 & \bf{2.82} & 2.93 & 3.40 \\
 & Sandwich & 4.30 & 3.82 & 3.41 & 3.04 & 2.90 & 2.87 & \bf{2.63} & 3.28 \\
 & Ours & \bf{4.27} & \bf{3.63} & \bf{3.04} & \bf{3.03} & \bf{2.68} & 2.90 & 2.82 & \bf{3.20} \\

\midrule
\midrule
 \multirow{3}{*}{AFHQv2 $64\times64$} & Uniform & 5.22 & 4.19 & 3.70 & 3.43 & 2.88 & 2.72 & 2.70 & 3.55 \\
  & Sandwich & 5.38 & 4.60 & 3.75 & 3.30 & 2.88 & \bf{2.31} & \bf{2.30} & 3.50 \\
 & Ours & \bf{5.04} & \bf{3.55} & \bf{2.94} & \bf{2.79} & \bf{2.46} & 2.43 & 2.37 & \bf{3.05}\\
\bottomrule
\end{tabular}
}
\label{tab:strategy}
\end{table*}

Next, we investigate the effect of our reweighting training strategy. Concretely, we compare our linear reweighting strategy clarified in Section \ref{sec:Greedy Allocation Strategy} with ablations: uniform weights on all subnetworks (Uniform) and Sandwich strategy (Sandwich) proposed in \citep{yu2019universally}: $w_{\cP_{1}} = w_{\cP_{N}} = 0.25$, and other $w_{\cP_{i}}$ are the same constant. 
\par
The comparison results conducted under U-Net are shown in Table~\ref{tab:strategy}. As can be seen, the uniform strategy performs the worst, especially at $0.25\times$, which illustrates the slow convergence of smaller subnetworks during OFA training. More details to the convergence rate of subnetworks in varied sizes refer to Appendix~C. 
In contrast, our reweighting strategy consistently outperforms the Uniform and Sandwich baselines across at almost rates over all datasets, showing that our strategy balances the optimization of subnetworks.

\begin{figure}[t]
    \small
    \centering
    \includegraphics[width=\columnwidth]{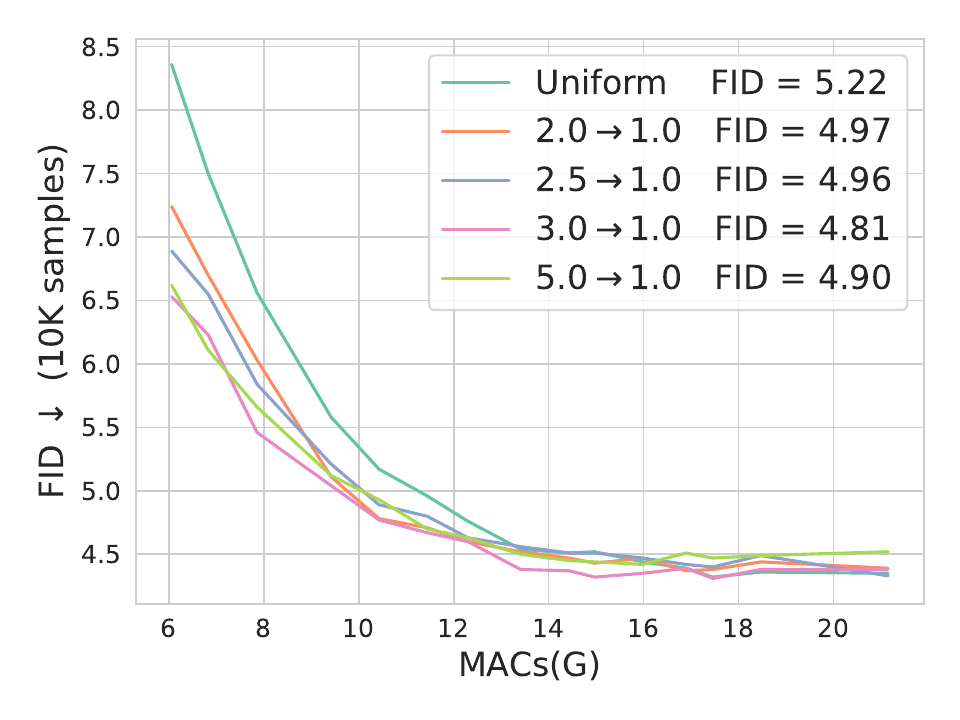}
    \caption{Comparison of FID versus MACs with different reweighting ratios on CIFAR10 under U-Net.
    $m\to 1.0$ means $w_{\cP_{1}} = mw_{\cP_{N}}$.
    We also report the averaged FID over subnetworks.
    For efficiency, we evaluate FID on 10K generated samples.}
    \label{figs:reweight_value_lineplot}
    \vspace{-0.15in}
\end{figure}

Furthermore, we vary the reweighting ratio between $w_{\cP_{1}}$ and $w_{\cP_{N}}$ in our reweighting strategy.
The results are shown in Figure~\ref{figs:reweight_value_lineplot}, where $m\to 1.0$ means $w_{\cP_{1}} = mw_{\cP_{N}}$. 
When the ratio $m$ is 3.0 ($3.0\to 1.0$), the averaged performance is better than the others.

\subsection{Discussion}\label{sec:discussion}
\paragraph{Update Importance during Training.}
In Section~\ref{sec:channel_importance}, 
we obtain the importance scores of the channels based on the pre-trained model and fix them in the whole training.
We also explore updating the importance score during OFA training to construct subnetworks.

\paragraph{Compute Importance for Different Time.}
As the DPMs can be recognized as multitask learning under different time steps $t$ \citep{balaji2022ediffi}, we also explore calculating the importance score of channels for different $t$.
That is, for the target $\cP_{i}$, the different subnetworks are extracted from the OFA network, which varied with $t$.

Unfortunately, compared to the methods in Algorithm \ref{alg: construction}, both variants obtain worse results.
We speculate that these variants implicitly increase the number of candidate subnetworks and may introduce noise in optimization, both of which can slow down the convergence of OFA training.
More details refer to the Appendix~E.

\section{Conclusion}
In this work, we propose a novel compression framework for DPMs: the \textit{once-for-all} (OFA) Diffusion Compression framework, which yields subnetworks of various computational configurations and can flexibly adapt to different hardware and user latency requirements.
To alleviate interference between subnetworks, we propose to construct subnetworks based on parameter importance, which identifies and shares important channels among subnetworks and allocates more parameters to more important layers. 
Furthermore, we propose a reweighting strategy in the training to balance the optimization processes for subnetworks of different sizes.
Experimental results on varied architectures show that for the same parameter size, the subnetworks compressed by our OFA framework not only consistently outperform the other OFA baselines but also achieve comparable or even better performance than separate compression, while enjoying a large reduction in training overhead.
Since our OFA-Diffusion Compression framework can reduce carbon emissions and is more environmentally friendly, we hope that the framework will facilitate the deployment of diffusion models on a variety of resource-constrained devices for real-life applications.

\bibliographystyle{ACM-Reference-Format}
\bibliography{reference}

\appendix

\clearpage
\section{OFA-Diffusion framework on U-Net}
\label{sec:ofa_unet}
\begin{figure*}[t]
\vspace{-0.1in}
    \begin{center}
    \includegraphics[width=0.95\linewidth]{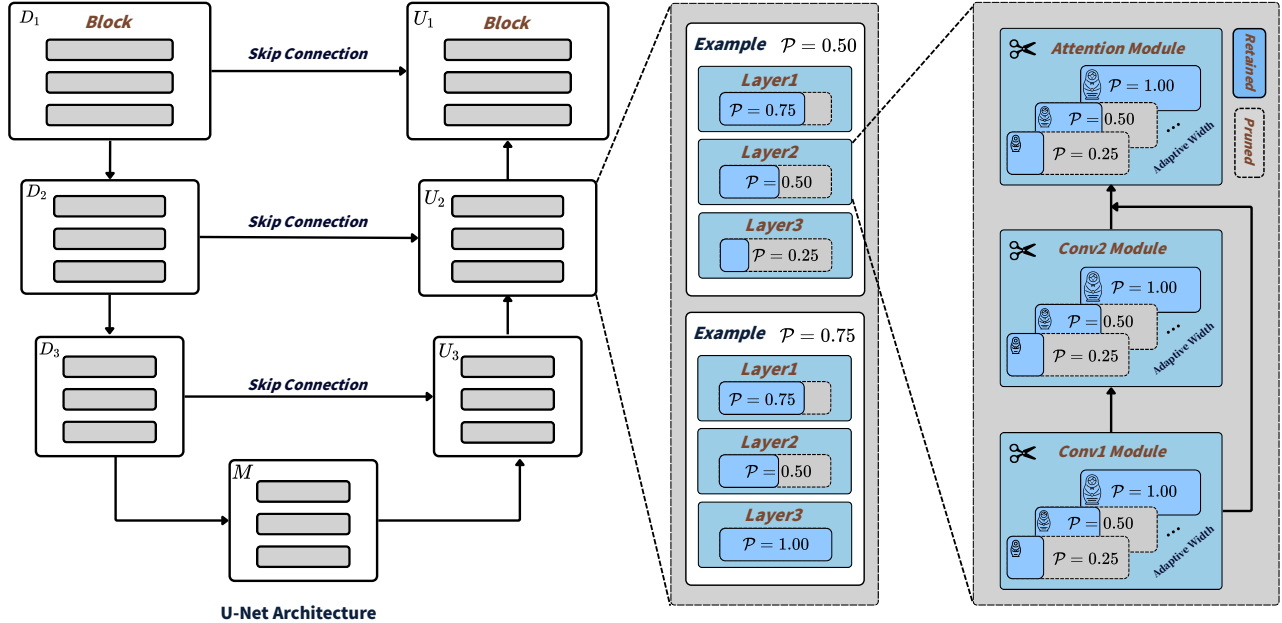}
    \caption{
    Overview of our OFA-Diffusion Compression framework.
    On the left, we present the structure of U-Net architecture.
    On the right, we show the details of layer in blocks and the compressed modules ($Conv_1$, $Conv_2$ and Self-Attention) in our compression framework. We omit GroupNorm and time embedding for simplicity.Given a retention rate $\cP_{i}$, only the top-$\cP_{i}$ important parameters will be preserved in each block, then these budgets will be unevenly allocated to each layer in a greedy manner. By doing so, for two retention rates $\cP_{i} < \cP_{j}$, in each layer, the corresponded module of $\cP_{i}$ is a subset of the module corresponded to $\cP_{j}$. 
    }
    \label{figs:overview_unet}
    \end{center}
\end{figure*}

As illustrated in Fig.~\ref{figs:overview_unet}, we present the overall pipeline of our OFA-Diffusion compression framework built upon the U-Net architecture.
\section{Hyperparameters}
\label{sec:hyperparameters}
We adopt the unconditional EDM~\citep{edm} VP model to initialize our OFA diffusion network in the CIFAR10\footnote{https://nvlabs-fi-cdn.nvidia.com/edm/pretrained/edm-cifar10-32x32-uncond-vp.pkl}, FFHQ\footnote{https://nvlabs-fi-cdn.nvidia.com/edm/pretrained/edm-ffhq-64x64-uncond-vp.pkl}, and AFHQv2\footnote{https://nvlabs-fi-cdn.nvidia.com/edm/pretrained/edm-afhqv2-64x64-uncond-vp.pkl} experiments.
Regarding the ImageNet dataset, we use the ADM architecture~\citep{dhariwal2021adm} with the EDM precondition\footnote{https://nvlabs-fi-cdn.nvidia.com/edm/pretrained/edm-imagenet-64x64-cond-adm.pkl}.
For U-ViT, we use the pretrained U-ViT-S/2 model on CIFAR10\footnote{\url{https://drive.google.com/file/d/1yoYyuzR_hQYWU0mkTj659tMTnoCWCMv-/view}} and the U-ViT-S/4 model on CelebA $64\times64$\footnote{\url{https://drive.google.com/file/d/13YpbRtlqF1HDBNLNRlKxLTbKbKeLE06C/view}} provided by Bao~et~al.~\citep{uvit}. For Stable Diffusion~\citep{sd}, we use the pretrained SD~v1.5\footnote{\url{https://huggingface.co/runwayml/stable-diffusion-v1-5}} released by RunwayML.


\begin{table*}[t]
\centering
\small
\caption{Hyperparameters of our OFA diffusion network on each dataset.}
\resizebox{\linewidth}{!}{
\begin{tabular}{l | c c c c | c c | c}
\toprule
 & \multicolumn{4}{c|}{EDM (UNet)} & \multicolumn{2}{c|}{UViT} & SD \\
\cmidrule(lr){2-5}\cmidrule(lr){6-7}\cmidrule(lr){8-8}
Hyperparameters
  & CIFAR10 $32\times32$ & FFHQ $64\times64$ & AFHQv2 $64\times64$ & ImageNet $64\times64$
  & CIFAR10 $32\times32$ & CelebA $64\times64$
  & COCO $512\times512$ \\
\midrule
Target retention rates $|\mathcal{P}|$ & 28 & 28 & 28 & 22 & 19 & 19 & 28 \\
Batch size         & 512  & 256  & 256  & 512  & 512  & 256  & 64   \\
Learning rate      & 1e-3 & 2e-4 & 2e-4 & 1e-4 & 2e-4 & 1e-4 & 5e-5 \\
Training iterations & 200K & 200K & 200K & 400K & 50K & 100K & 50K  \\
EMA half-life (Mimg) & 0.5 & 0.5 & 0.5 & 50 & 3.5 & 1.8 & 0.4    \\
Augment probability & 0.12 & 0.15 & 0.15 & 0.0 & 0.5   & 0.5   & 0.0    \\
Dropout            & 0.13 & 0.05 & 0.25 & 0.1  & 0    & 0    & 0    \\
Sampling NFE       & 35 &\multicolumn{3}{c|}{79 (Deterministic Heun's \citep{edm})}
                   & \multicolumn{2}{c|}{50 (DPM-Solver)} & 50 (DDIM) \\
\bottomrule
\end{tabular}
}
\label{tab:hyperparameters}
\end{table*}


For optimization, we strictly adhere to the default settings of the respective official training scripts to ensure fair comparisons. Specifically, we employ Adam~\citep{adam} for the EDM architecture, and AdamW~\citep{adamw} for both U-ViT and Stable Diffusion, preserving their original $\beta$ configurations. Furthermore, to stabilize the generation quality, we consistently maintain an exponential moving average (EMA) of the network weights. 

Detailed hyperparameters are shown in Table~\ref{tab:hyperparameters}.

\section{The Convergence of Different-Sized Subnetworks}
\label{app:convergence_with_size}

\begin{figure}[t]
    \centering
    \includegraphics[width=0.48\textwidth]
    {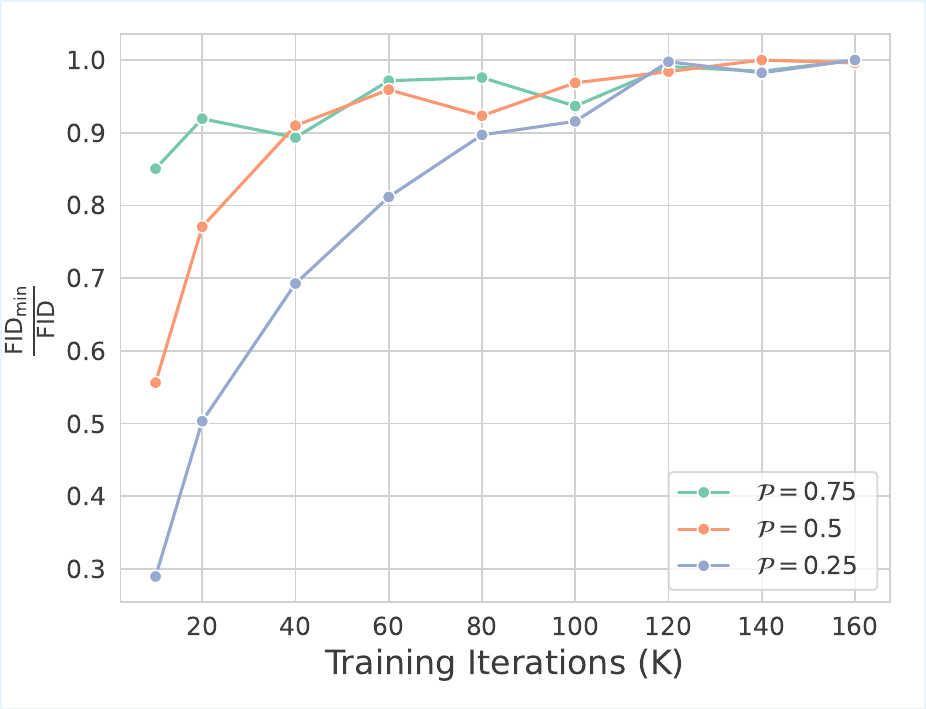}
    \caption{The convergence of subnetworks with different parameter retention rates $\cP$ on dataset CIFAR10 $32\times32$. The y-axis is the relative FID scores, to indicate the convergence rate under different retention rates. 
    }
    \label{fig:cifar10_train_duration}
\end{figure}

In this section, we show the convergence of subnetworks with different retention rates $\cP$ for the U-Net architecture on the CIFAR10 $32\times32$ dataset. 
The subnetworks in the OFA-Diffusion Compression framework are initialized from the pre-trained diffusion model. 
Intuitively, larger subnetworks have more capacity, leading to faster convergence, while smaller ones converge more slowly. 
We report the ratio of the minimum value of FID during training to the FID at different iterations in Figure~\ref{fig:cifar10_train_duration}. 
We observe that the largest subnetwork ($\cP = 0.75$) has the fastest convergence, while the smallest subnetwork ($\cP = 0.25$) has the slowest convergence.

\section{Visualization of Layer Importance}\label{app: importance scores in layer}

\begin{figure}[h]
    \centering
    \includegraphics[width=\linewidth]{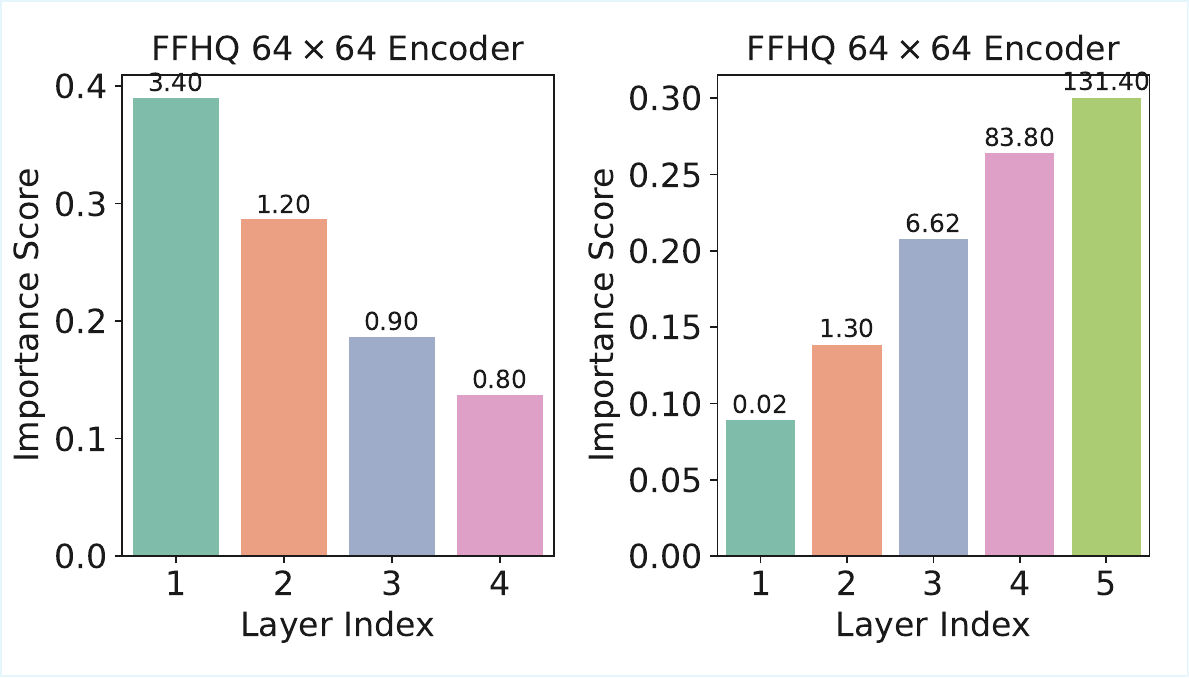}
    
    \vspace{1em} 
    
    \includegraphics[width=\linewidth]{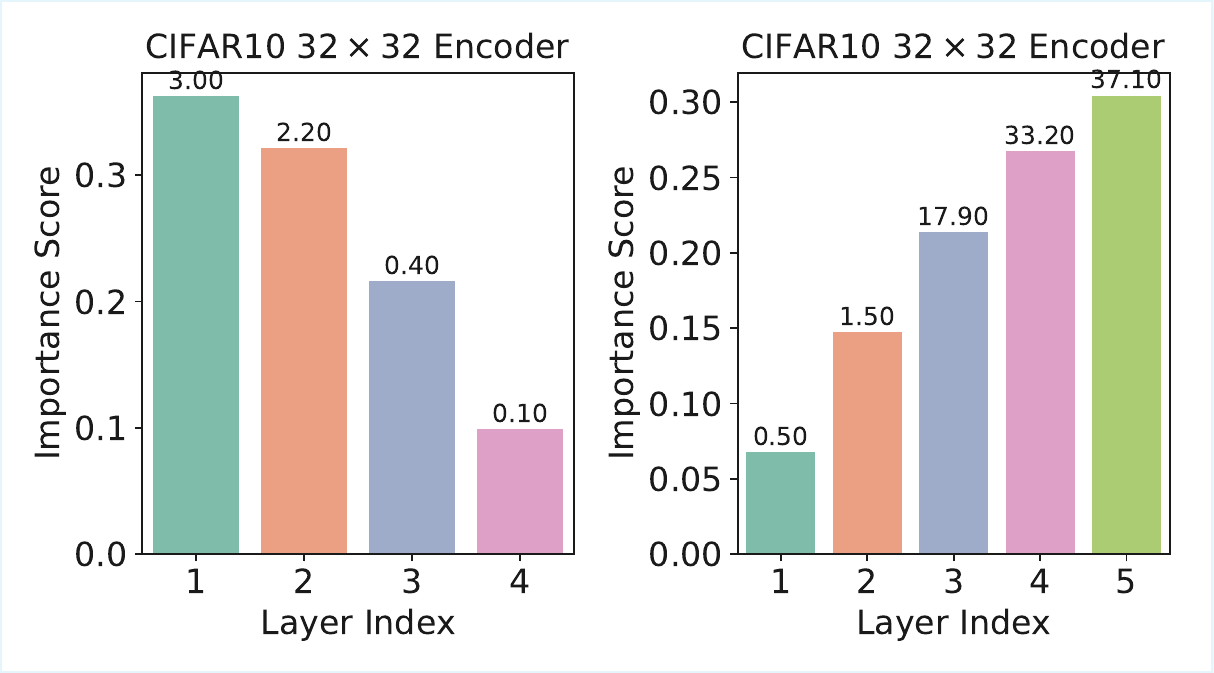}
    
    \caption{Visualization of normed importance scores across layers in the block at $8\times8$ resolution.
    At the top of each bar, we also present the FID increases (denoting performance drop) if we remove the corresponding layer, which shows the sensitivity of layers.}
    \label{fig: layer_importance}
\end{figure}

In this section, we attempt to visualize the varying importance of layers (calculated as in Section~\ref{sec:Greedy Allocation Strategy}) in the diffusion model.
In Figure~\ref{fig: layer_importance}, we visualize layer importance in the $8\times8$ resolution block of the pre-trained EDM~\citep{edm} on CIFAR10 $32\times32$ and FFHQ $64\times64$.
We also present the sensitivity of layers, which is the FID increase (denoting the performance drop) after removing the corresponding layer.
A significant decrease in performance indicates the important role of that layer in the DPM~\citep{bk-sdm}.

From Fig.~\ref{fig: layer_importance}, we can observe that the importance of layers is consistent with their sensitivity. 
We find that the innermost layers (the 4th layer of the encoder and the 1st layer of the decoder) have the least importance.


\begin{table}[t]
\caption{
Comparison of varying importance scores of channels as discussed in Section~\ref{app:update importance scores}.
For efficiency, we evaluate FID on 10K generated samples.
} 
\small\centering
\begin{tabular}{c|cccc}
\toprule
Method & $0.25\times$ & $0.5\times$ & $0.75\times$ & $1.0\times$ \\
\midrule
 Ours (Algorithm \ref{alg: construction}) & 5.80 & 4.73 & 4.37 & 4.20 \\
 (a) Update Importance Scores& 9.52 & 5.92 & 4.44 & 4.39 \\
 (b) Split Time & 8.00 & 5.84 & 4.87 & 4.34 \\

\bottomrule
\end{tabular}
\label{tab:update_and_split}
\end{table}

\section{Discussions on Importance Scores}\label{app:update importance scores}

As clarified in Section \ref{sec:discussion}, channel importance scores can (a) be updated during training or (b) be calculated at different times $t$.
We experiment with both variants in subnetwork construction:
\par
\begin{enumerate}
    \item For variant (a), we update the importance scores of channels after every epoch of training, where the importance scores are computed with the sensitivity metric, the same as in \eqref{eq: compute importance}. 
    \item Since diffusion training can be viewed as multitask learning over different time steps $t$ \citep{balaji2022ediffi}, the importance scores can be varied over different $t$. 
    Instead of computing importance scores over the whole time interval in EDM ($t\in [\sigma^{-1}(0.002), \allowbreak \sigma^{-1}(80.0)]$\footnote{$\sigma^{-1}$ denotes the inversion function of $\sigma$.}), the variant (b) splits the time into 3 intervals: $t_1\in [\sigma^{-1}(0.002), \allowbreak \sigma^{-1}(0.1)]$, $t_2\in [\sigma^{-1}(0.1), \allowbreak \sigma^{-1}(1.0)]$, and $t_3\in [\sigma^{-1}(1.0), \allowbreak \sigma^{-1}(80.0)]$. 
    For each target $\cP_{i}$, the variant (b) constructs three different subnetworks and computes their importance scores according to their corresponding $t$. During the sampling process under a specific $\cP_{i}$, we use one of the three corresponding models, depending on the time $t$.
\end{enumerate}
We compare the above two variants (a) and (b) with the construction approach in Section~\ref{sec:construct_subnetwork} on CIFAR10 $32\times32$. 
For simplicity, OFA networks are targeted at 4 parameter retention rates $\cP_{i} \in \left[0.25, 0.5, 0.75, 1.0 \right]$ and trained for $100$K iterations.
The results are shown in Table~\ref{tab:update_and_split}.

\par
As can be seen, compared with our method in Algorithm \ref{alg: construction}, both variants obtain worse results.
Thus, we speculate that the two variants are not wise choices in practice. 
We speculate that these variants implicitly increase the number of candidate subnetworks and may introduce noise in OFA network optimization, both of which can slow down the convergence of training.

\section{Samples of Different Subnetworks}
\label{app:visualization}
\begin{figure*}[!h]
    \centering
    \setlength{\tabcolsep}{1pt}
    \resizebox{1.0\linewidth}{!}{
    \begin{tabular}{c c c c c c c c}
     & $\cP=0.25\times$ & $\cP=0.375\times$ & $\cP=0.5\times$ & $\cP=0.625\times$ & $\cP=0.75\times$ & $\cP=0.875\times$ & $\cP=1.0\times$\\
    \raisebox{2.0\height}{\makecell{Separate Compression \\ Avg. FID = 2.49}} &
    \begin{subfigure}{0.15\textwidth}
        \includegraphics[width=\textwidth]{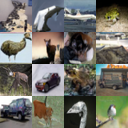}
    \end{subfigure} &
    \begin{subfigure}{0.15\textwidth}
        \includegraphics[width=\textwidth]{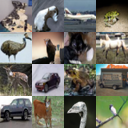}
    \end{subfigure} &
    \begin{subfigure}{0.15\textwidth}
        \includegraphics[width=\textwidth]{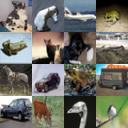}
    \end{subfigure} &
    \begin{subfigure}{0.15\textwidth}
        \includegraphics[width=\textwidth]{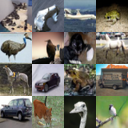}
    \end{subfigure} &
        \begin{subfigure}{0.15\textwidth}
        \includegraphics[width=\textwidth]{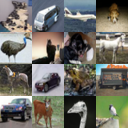}
    \end{subfigure} &
        \begin{subfigure}{0.15\textwidth}
        \includegraphics[width=\textwidth]{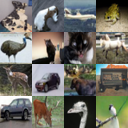}
    \end{subfigure} &
        \begin{subfigure}{0.15\textwidth}
        \includegraphics[width=\textwidth]{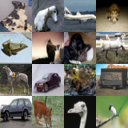}
    \end{subfigure} \\
    \raisebox{2.0\height}{\makecell{OFA Diffusion Compression \\ Avg. FID = 2.75}} &
    \begin{subfigure}{0.15\textwidth}
        \includegraphics[width=\textwidth]{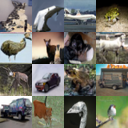}
    \end{subfigure} &
    \begin{subfigure}{0.15\textwidth}
        \includegraphics[width=\textwidth]{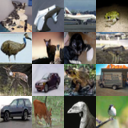}
    \end{subfigure} &
    \begin{subfigure}{0.15\textwidth}
        \includegraphics[width=\textwidth]{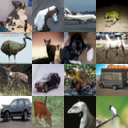}
    \end{subfigure} &
    \begin{subfigure}{0.15\textwidth}
        \includegraphics[width=\textwidth]{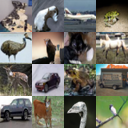}
    \end{subfigure} &
        \begin{subfigure}{0.15\textwidth}
        \includegraphics[width=\textwidth]{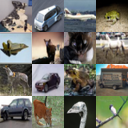}
    \end{subfigure} &
        \begin{subfigure}{0.15\textwidth}
        \includegraphics[width=\textwidth]{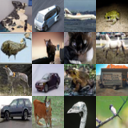}
    \end{subfigure} &
        \begin{subfigure}{0.15\textwidth}
        \includegraphics[width=\textwidth]{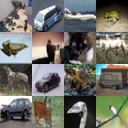}
    \end{subfigure} \\
    \raisebox{1.8 \height}{\makecell{OFA Diffusion Compression \\ +Fine-tune 10K \\ Avg. FID = 2.41}} &
    \begin{subfigure}{0.15\textwidth}
        \includegraphics[width=\textwidth]{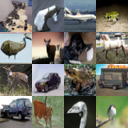}
    \end{subfigure} &
    \begin{subfigure}{0.15\textwidth}
        \includegraphics[width=\textwidth]{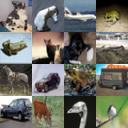}
    \end{subfigure} &
    \begin{subfigure}{0.15\textwidth}
        \includegraphics[width=\textwidth]{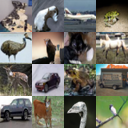}
    \end{subfigure} &
    \begin{subfigure}{0.15\textwidth}
        \includegraphics[width=\textwidth]{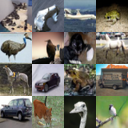}
    \end{subfigure} &
        \begin{subfigure}{0.15\textwidth}
        \includegraphics[width=\textwidth]{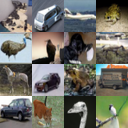}
    \end{subfigure} &
        \begin{subfigure}{0.15\textwidth}
        \includegraphics[width=\textwidth]{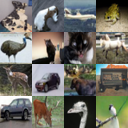}
    \end{subfigure} &
        \begin{subfigure}{0.15\textwidth}
        \includegraphics[width=\textwidth]{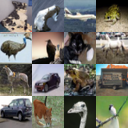}
    \end{subfigure} \\
    \raisebox{1.8\height}{\makecell{OFA Diffusion Compression \\ + Fine-tune 20K \\ Avg. FID = 2.29}} &
    \begin{subfigure}{0.15\textwidth}
        \includegraphics[width=\textwidth]{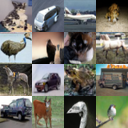}
    \end{subfigure} &
    \begin{subfigure}{0.15\textwidth}
        \includegraphics[width=\textwidth]{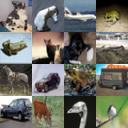}
    \end{subfigure} &
    \begin{subfigure}{0.15\textwidth}
        \includegraphics[width=\textwidth]{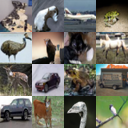}
    \end{subfigure} &
    \begin{subfigure}{0.15\textwidth}
        \includegraphics[width=\textwidth]{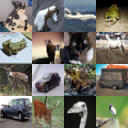}
    \end{subfigure} &
        \begin{subfigure}{0.15\textwidth}
        \includegraphics[width=\textwidth]{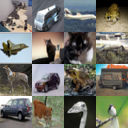}
    \end{subfigure} &
        \begin{subfigure}{0.15\textwidth}
        \includegraphics[width=\textwidth]{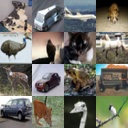}
    \end{subfigure} &
        \begin{subfigure}{0.15\textwidth}
        \includegraphics[width=\textwidth]{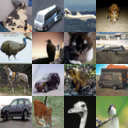}
    \end{subfigure} \\
    \end{tabular}
    }
    \caption{Samples from subnetworks trained by different approaches with the same random seed on CIFAR10 $32\times32$ (EDM unconditional sampling, U-Net backbone).}
\label{fig: vis_cifar10}
\end{figure*}

\begin{figure*}[!htb]
    \centering
    \setlength{\tabcolsep}{1pt}
    \resizebox{1.0\linewidth}{!}{
    \begin{tabular}{c c c c c c c c}
     & $\cP=0.25\times$ & $\cP=0.375\times$ & $\cP=0.5\times$ & $\cP=0.625\times$ & $\cP=0.75\times$ & $\cP=0.875\times$ & $\cP=1.0\times$\\
    \raisebox{2.0\height}{\makecell{Separate Compression \\ Avg. FID = 3.17}} &
    \begin{subfigure}{0.15\textwidth}
        \includegraphics[width=\textwidth]{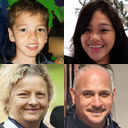}
    \end{subfigure} &
    \begin{subfigure}{0.15\textwidth}
        \includegraphics[width=\textwidth]{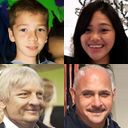}
    \end{subfigure} &
    \begin{subfigure}{0.15\textwidth}
        \includegraphics[width=\textwidth]{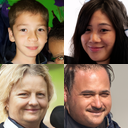}
    \end{subfigure} &
    \begin{subfigure}{0.15\textwidth}
        \includegraphics[width=\textwidth]{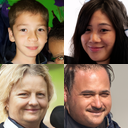}
    \end{subfigure} &
        \begin{subfigure}{0.15\textwidth}
        \includegraphics[width=\textwidth]{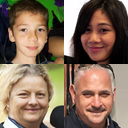}
    \end{subfigure} &
        \begin{subfigure}{0.15\textwidth}
        \includegraphics[width=\textwidth]{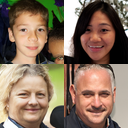}
    \end{subfigure} &
        \begin{subfigure}{0.15\textwidth}
        \includegraphics[width=\textwidth]{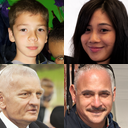}
    \end{subfigure} \\
    \raisebox{2.0\height}{\makecell{OFA Diffusion Compression \\ Avg. FID = 3.24}} &
    \begin{subfigure}{0.15\textwidth}
        \includegraphics[width=\textwidth]{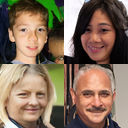}
    \end{subfigure} &
    \begin{subfigure}{0.15\textwidth}
        \includegraphics[width=\textwidth]{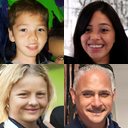}
    \end{subfigure} &
    \begin{subfigure}{0.15\textwidth}
        \includegraphics[width=\textwidth]{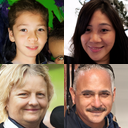}
    \end{subfigure} &
    \begin{subfigure}{0.15\textwidth}
        \includegraphics[width=\textwidth]{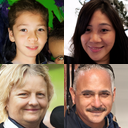}
    \end{subfigure} &
        \begin{subfigure}{0.15\textwidth}
        \includegraphics[width=\textwidth]{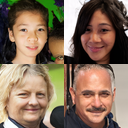}
    \end{subfigure} &
        \begin{subfigure}{0.15\textwidth}
        \includegraphics[width=\textwidth]{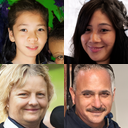}
    \end{subfigure} &
        \begin{subfigure}{0.15\textwidth}
        \includegraphics[width=\textwidth]{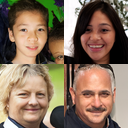}
    \end{subfigure} \\
    \raisebox{1.8 \height}{\makecell{OFA Diffusion Compression \\ +Fine-tune 10K \\ Avg. FID = 3.06}} &
    \begin{subfigure}{0.15\textwidth}
        \includegraphics[width=\textwidth]{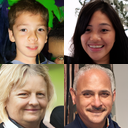}
    \end{subfigure} &
    \begin{subfigure}{0.15\textwidth}
        \includegraphics[width=\textwidth]{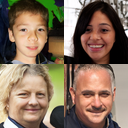}
    \end{subfigure} &
    \begin{subfigure}{0.15\textwidth}
        \includegraphics[width=\textwidth]{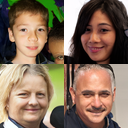}
    \end{subfigure} &
    \begin{subfigure}{0.15\textwidth}
        \includegraphics[width=\textwidth]{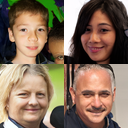}
    \end{subfigure} &
        \begin{subfigure}{0.15\textwidth}
        \includegraphics[width=\textwidth]{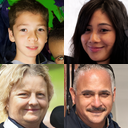}
    \end{subfigure} &
        \begin{subfigure}{0.15\textwidth}
        \includegraphics[width=\textwidth]{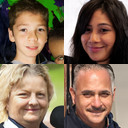}
    \end{subfigure} &
        \begin{subfigure}{0.15\textwidth}
        \includegraphics[width=\textwidth]{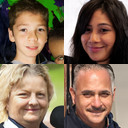}
    \end{subfigure} \\
    \raisebox{1.8\height}{\makecell{OFA Diffusion Compression \\ + Fine-tune 20K \\ Avg. FID = 3.04}} &
    \begin{subfigure}{0.15\textwidth}
        \includegraphics[width=\textwidth]{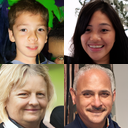}
    \end{subfigure} &
    \begin{subfigure}{0.15\textwidth}
        \includegraphics[width=\textwidth]{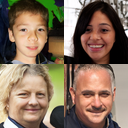}
    \end{subfigure} &
    \begin{subfigure}{0.15\textwidth}
        \includegraphics[width=\textwidth]{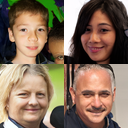}
    \end{subfigure} &
    \begin{subfigure}{0.15\textwidth}
        \includegraphics[width=\textwidth]{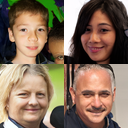}
    \end{subfigure} &
        \begin{subfigure}{0.15\textwidth}
        \includegraphics[width=\textwidth]{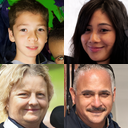}
    \end{subfigure} &
        \begin{subfigure}{0.15\textwidth}
        \includegraphics[width=\textwidth]{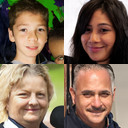}
    \end{subfigure} &
        \begin{subfigure}{0.15\textwidth}
        \includegraphics[width=\textwidth]{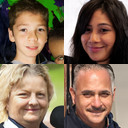}
    \end{subfigure} \\
    \end{tabular}
    }
    \caption{Samples from subnetworks trained by different approaches with the same random seed on FFHQ $64\times64$ (EDM unconditional sampling, U-Net backbone).}
\label{fig: vis_ffhq}
\end{figure*}

\begin{figure*}[!htb]
    \centering
    \setlength{\tabcolsep}{1pt}
    \resizebox{1.0\linewidth}{!}{
    \begin{tabular}{c c c c c c c c}
     & $\cP=0.25\times$ & $\cP=0.375\times$ & $\cP=0.5\times$ & $\cP=0.625\times$ & $\cP=0.75\times$ & $\cP=0.875\times$ & $\cP=1.0\times$\\
    \raisebox{2.0\height}{\makecell{Separate Compression \\ Avg. FID = 3.32}} &
    \begin{subfigure}{0.15\textwidth}
        \includegraphics[width=\textwidth]{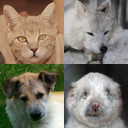}
    \end{subfigure} &
    \begin{subfigure}{0.15\textwidth}
        \includegraphics[width=\textwidth]{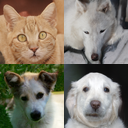}
    \end{subfigure} &
    \begin{subfigure}{0.15\textwidth}
        \includegraphics[width=\textwidth]{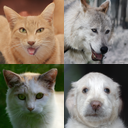}
    \end{subfigure} &
    \begin{subfigure}{0.15\textwidth}
        \includegraphics[width=\textwidth]{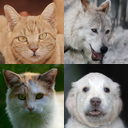}
    \end{subfigure} &
        \begin{subfigure}{0.15\textwidth}
        \includegraphics[width=\textwidth]{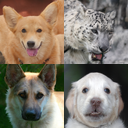}
    \end{subfigure} &
        \begin{subfigure}{0.15\textwidth}
        \includegraphics[width=\textwidth]{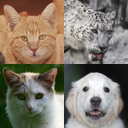}
    \end{subfigure} &
        \begin{subfigure}{0.15\textwidth}
        \includegraphics[width=\textwidth]{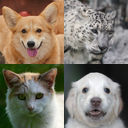}
    \end{subfigure} \\
    \raisebox{2.0\height}{\makecell{OFA Diffusion Compression \\ Avg. FID = 3.05}} &
    \begin{subfigure}{0.15\textwidth}
        \includegraphics[width=\textwidth]{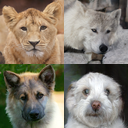}
    \end{subfigure} &
    \begin{subfigure}{0.15\textwidth}
        \includegraphics[width=\textwidth]{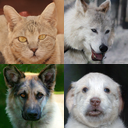}
    \end{subfigure} &
    \begin{subfigure}{0.15\textwidth}
        \includegraphics[width=\textwidth]{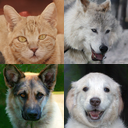}
    \end{subfigure} &
    \begin{subfigure}{0.15\textwidth}
        \includegraphics[width=\textwidth]{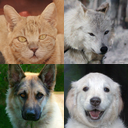}
    \end{subfigure} &
        \begin{subfigure}{0.15\textwidth}
        \includegraphics[width=\textwidth]{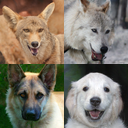}
    \end{subfigure} &
        \begin{subfigure}{0.15\textwidth}
        \includegraphics[width=\textwidth]{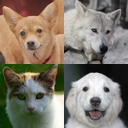}
    \end{subfigure} &
        \begin{subfigure}{0.15\textwidth}
        \includegraphics[width=\textwidth]{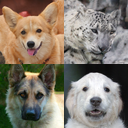}
    \end{subfigure} \\
    \raisebox{1.8 \height}{\makecell{OFA Diffusion Compression \\ +Fine-tune 10K \\ Avg. FID = 3.07}} &
    \begin{subfigure}{0.15\textwidth}
        \includegraphics[width=\textwidth]{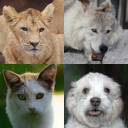}
    \end{subfigure} &
    \begin{subfigure}{0.15\textwidth}
        \includegraphics[width=\textwidth]{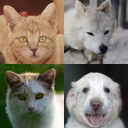}
    \end{subfigure} &
    \begin{subfigure}{0.15\textwidth}
        \includegraphics[width=\textwidth]{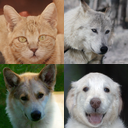}
    \end{subfigure} &
    \begin{subfigure}{0.15\textwidth}
        \includegraphics[width=\textwidth]{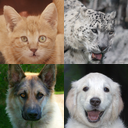}
    \end{subfigure} &
        \begin{subfigure}{0.15\textwidth}
        \includegraphics[width=\textwidth]{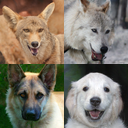}
    \end{subfigure} &
        \begin{subfigure}{0.15\textwidth}
        \includegraphics[width=\textwidth]{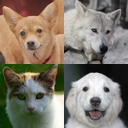}
    \end{subfigure} &
        \begin{subfigure}{0.15\textwidth}
        \includegraphics[width=\textwidth]{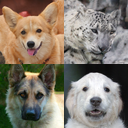}
    \end{subfigure} \\
    \raisebox{1.8\height}{\makecell{OFA Diffusion Compression \\ + Fine-tune 20K \\ Avg. FID = 3.00}} &
    \begin{subfigure}{0.15\textwidth}
        \includegraphics[width=\textwidth]{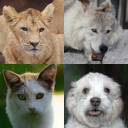}
    \end{subfigure} &
    \begin{subfigure}{0.15\textwidth}
        \includegraphics[width=\textwidth]{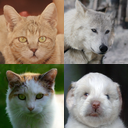}
    \end{subfigure} &
    \begin{subfigure}{0.15\textwidth}
        \includegraphics[width=\textwidth]{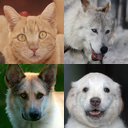}
    \end{subfigure} &
    \begin{subfigure}{0.15\textwidth}
        \includegraphics[width=\textwidth]{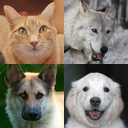}
    \end{subfigure} &
        \begin{subfigure}{0.15\textwidth}
        \includegraphics[width=\textwidth]{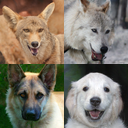}
    \end{subfigure} &
        \begin{subfigure}{0.15\textwidth}
        \includegraphics[width=\textwidth]{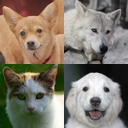}
    \end{subfigure} &
        \begin{subfigure}{0.15\textwidth}
        \includegraphics[width=\textwidth]{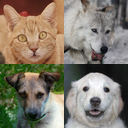}
    \end{subfigure} \\
    \end{tabular}
    }
    \caption{Samples from subnetworks trained by different approaches with the same random seed on AFHQv2 $64\times64$ (EDM unconditional sampling).}
\label{fig: vis_afhqv2}
\end{figure*}

\begin{figure*}[!htb]
    \centering
    \setlength{\tabcolsep}{1pt}
    \resizebox{1.0\linewidth}{!}{
    \begin{tabular}{c c c c c c c c}
     & $\cP=0.25\times$ & $\cP=0.375\times$ & $\cP=0.5\times$ & $\cP=0.625\times$ & $\cP=0.75\times$ & $\cP=0.875\times$ & $\cP=1.0\times$\\
    \raisebox{2.0\height}{\makecell{Separate Compression \\ Avg. FID = 6.08}} &
    \begin{subfigure}{0.15\textwidth}
        \includegraphics[width=\textwidth]{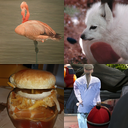}
    \end{subfigure} &
    \begin{subfigure}{0.15\textwidth}
        \includegraphics[width=\textwidth]{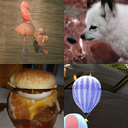}
    \end{subfigure} &
    \begin{subfigure}{0.15\textwidth}
        \includegraphics[width=\textwidth]{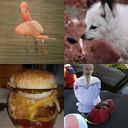}
    \end{subfigure} &
    \begin{subfigure}{0.15\textwidth}
        \includegraphics[width=\textwidth]{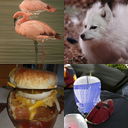}
    \end{subfigure} &
        \begin{subfigure}{0.15\textwidth}
        \includegraphics[width=\textwidth]{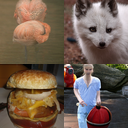}
    \end{subfigure} &
        \begin{subfigure}{0.15\textwidth}
        \includegraphics[width=\textwidth]{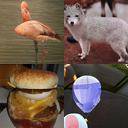}
    \end{subfigure} &
        \begin{subfigure}{0.15\textwidth}
        \includegraphics[width=\textwidth]{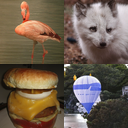}
    \end{subfigure} \\
    \raisebox{2.0\height}{\makecell{OFA Diffusion Compression \\ Avg. FID = 7.43}} &
    \begin{subfigure}{0.15\textwidth}
        \includegraphics[width=\textwidth]{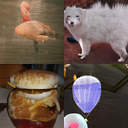}
    \end{subfigure} &
    \begin{subfigure}{0.15\textwidth}
        \includegraphics[width=\textwidth]{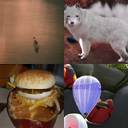}
    \end{subfigure} &
    \begin{subfigure}{0.15\textwidth}
        \includegraphics[width=\textwidth]{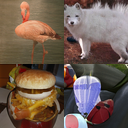}
    \end{subfigure} &
    \begin{subfigure}{0.15\textwidth}
        \includegraphics[width=\textwidth]{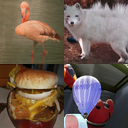}
    \end{subfigure} &
        \begin{subfigure}{0.15\textwidth}
        \includegraphics[width=\textwidth]{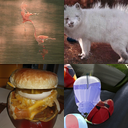}
    \end{subfigure} &
        \begin{subfigure}{0.15\textwidth}
        \includegraphics[width=\textwidth]{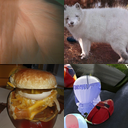}
    \end{subfigure} &
        \begin{subfigure}{0.15\textwidth}
        \includegraphics[width=\textwidth]{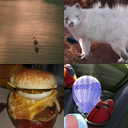}
    \end{subfigure} \\
    \raisebox{1.8 \height}{\makecell{OFA Diffusion Compression \\ +Fine-tune 20K \\ Avg. FID = 6.05}} &
    \begin{subfigure}{0.15\textwidth}
        \includegraphics[width=\textwidth]{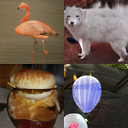}
    \end{subfigure} &
    \begin{subfigure}{0.15\textwidth}
        \includegraphics[width=\textwidth]{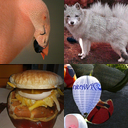}
    \end{subfigure} &
    \begin{subfigure}{0.15\textwidth}
        \includegraphics[width=\textwidth]{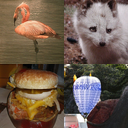}
    \end{subfigure} &
    \begin{subfigure}{0.15\textwidth}
        \includegraphics[width=\textwidth]{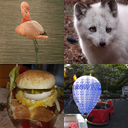}
    \end{subfigure} &
        \begin{subfigure}{0.15\textwidth}
        \includegraphics[width=\textwidth]{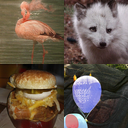}
    \end{subfigure} &
        \begin{subfigure}{0.15\textwidth}
        \includegraphics[width=\textwidth]{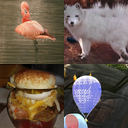}
    \end{subfigure} &
        \begin{subfigure}{0.15\textwidth}
        \includegraphics[width=\textwidth]{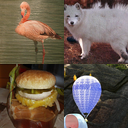}
    \end{subfigure} \\
    \raisebox{1.8\height}{\makecell{OFA Diffusion Compression \\ + Fine-tune 40K \\ Avg. FID = 5.47}} &
    \begin{subfigure}{0.15\textwidth}
        \includegraphics[width=\textwidth]{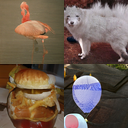}
    \end{subfigure} &
    \begin{subfigure}{0.15\textwidth}
        \includegraphics[width=\textwidth]{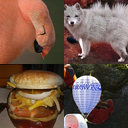}
    \end{subfigure} &
    \begin{subfigure}{0.15\textwidth}
        \includegraphics[width=\textwidth]{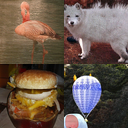}
    \end{subfigure} &
    \begin{subfigure}{0.15\textwidth}
        \includegraphics[width=\textwidth]{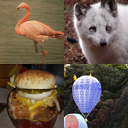}
    \end{subfigure} &
        \begin{subfigure}{0.15\textwidth}
        \includegraphics[width=\textwidth]{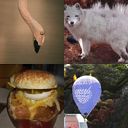}
    \end{subfigure} &
        \begin{subfigure}{0.15\textwidth}
        \includegraphics[width=\textwidth]{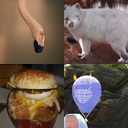}
    \end{subfigure} &
        \begin{subfigure}{0.15\textwidth}
        \includegraphics[width=\textwidth]{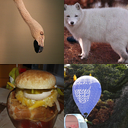}
    \end{subfigure} \\
    \end{tabular}
    }
    \caption{Samples from subnetworks trained by different approaches with the same random seed on ImageNet $64\times64$ (EDM conditional sampling, U-Net backbone).}
\label{fig: vis_imagenet}
\end{figure*}

\begin{figure*}[!h]
    \centering
    \setlength{\tabcolsep}{1pt}
    \resizebox{1.0\linewidth}{!}{
    \begin{tabular}{c c c c c c c c}
     & $\cP=0.25\times$ & $\cP=0.375\times$ & $\cP=0.5\times$ & $\cP=0.625\times$ & $\cP=0.75\times$ & $\cP=0.875\times$ & $\cP=1.0\times$\\
    \raisebox{2.0\height}{\makecell{Separate Compression \\ Avg. FID = 5.43}} &
    \begin{subfigure}{0.15\textwidth}
        \includegraphics[width=\textwidth]{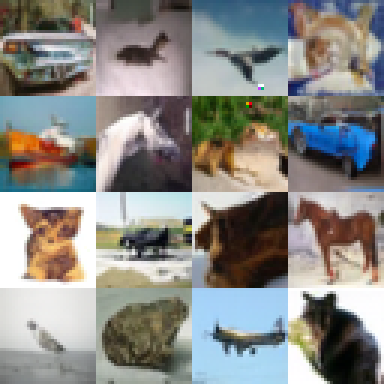}
    \end{subfigure} &
    \begin{subfigure}{0.15\textwidth}
        \includegraphics[width=\textwidth]{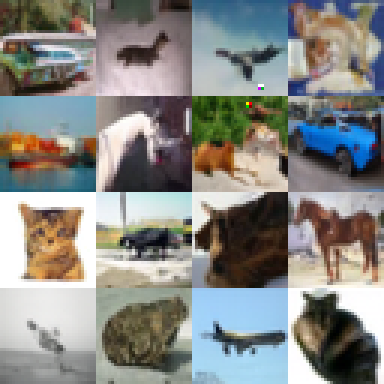}
    \end{subfigure} &
    \begin{subfigure}{0.15\textwidth}
        \includegraphics[width=\textwidth]{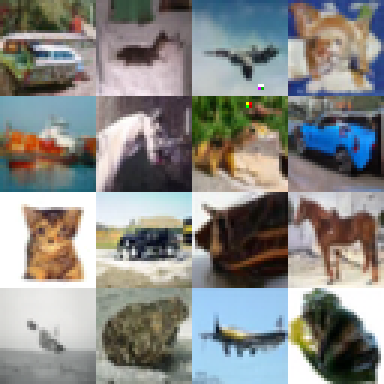}
    \end{subfigure} &
    \begin{subfigure}{0.15\textwidth}
        \includegraphics[width=\textwidth]{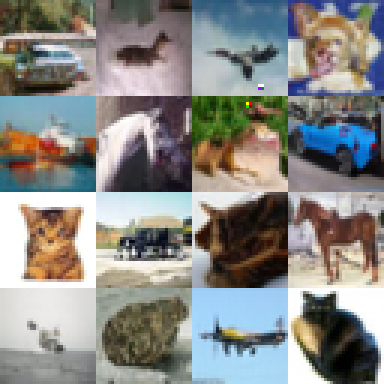}
    \end{subfigure} &
        \begin{subfigure}{0.15\textwidth}
        \includegraphics[width=\textwidth]{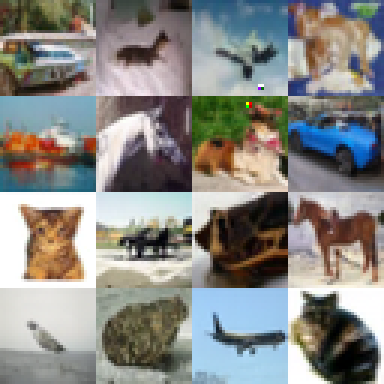}
    \end{subfigure} &
        \begin{subfigure}{0.15\textwidth}
        \includegraphics[width=\textwidth]{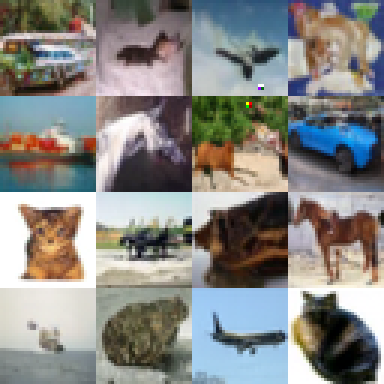}
    \end{subfigure} &
        \begin{subfigure}{0.15\textwidth}
        \includegraphics[width=\textwidth]{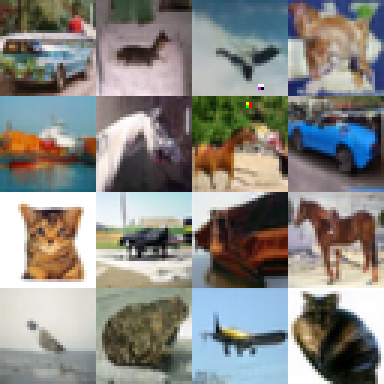}
    \end{subfigure} \\
    \raisebox{2.0\height}{\makecell{OFA Diffusion Compression \\ Avg. FID = 8.20}} &
    \begin{subfigure}{0.15\textwidth}
        \includegraphics[width=\textwidth]{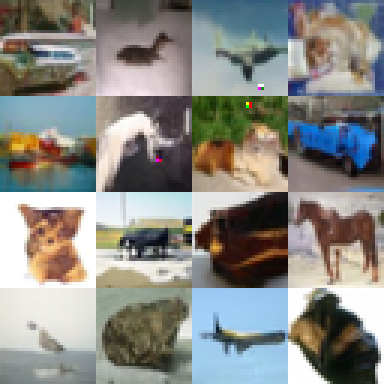}
    \end{subfigure} &
    \begin{subfigure}{0.15\textwidth}
        \includegraphics[width=\textwidth]{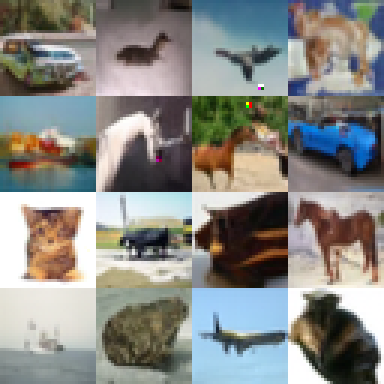}
    \end{subfigure} &
    \begin{subfigure}{0.15\textwidth}
        \includegraphics[width=\textwidth]{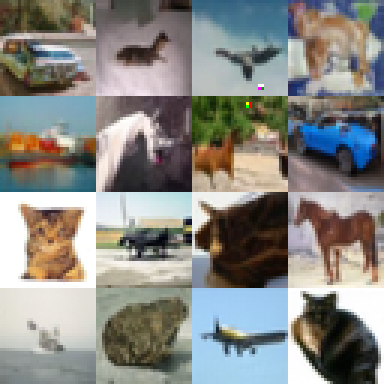}
    \end{subfigure} &
    \begin{subfigure}{0.15\textwidth}
        \includegraphics[width=\textwidth]{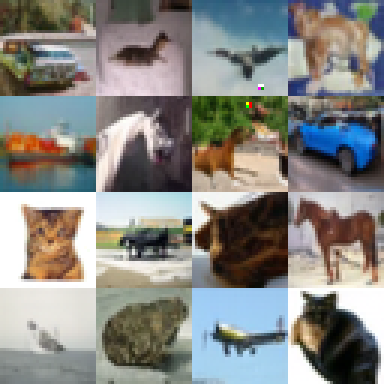}
    \end{subfigure} &
        \begin{subfigure}{0.15\textwidth}
        \includegraphics[width=\textwidth]{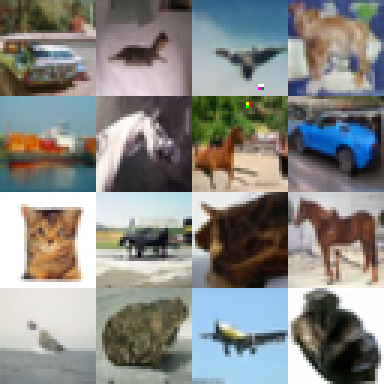}
    \end{subfigure} &
        \begin{subfigure}{0.15\textwidth}
        \includegraphics[width=\textwidth]{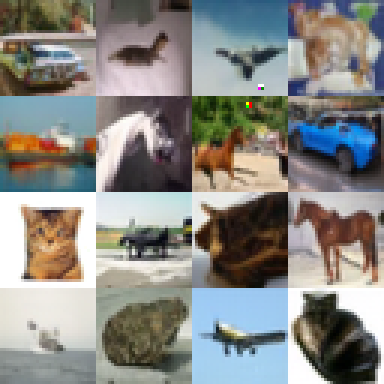}
    \end{subfigure} &
        \begin{subfigure}{0.15\textwidth}
        \includegraphics[width=\textwidth]{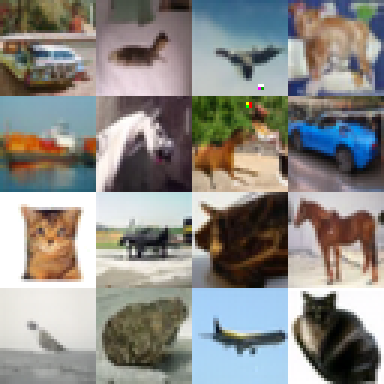}
    \end{subfigure} \\
    \raisebox{1.8 \height}{\makecell{OFA Diffusion Compression \\ +Fine-tune 10K \\ Avg. FID = 6.91}} &
    \begin{subfigure}{0.15\textwidth}
        \includegraphics[width=\textwidth]{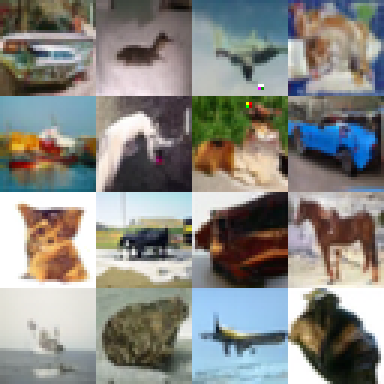}
    \end{subfigure} &
    \begin{subfigure}{0.15\textwidth}
        \includegraphics[width=\textwidth]{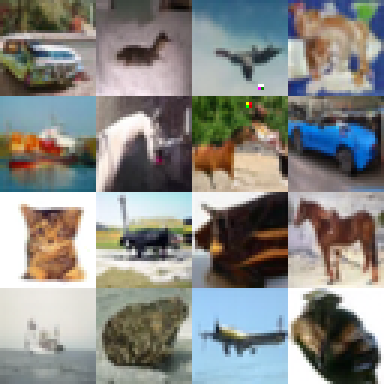}
    \end{subfigure} &
    \begin{subfigure}{0.15\textwidth}
        \includegraphics[width=\textwidth]{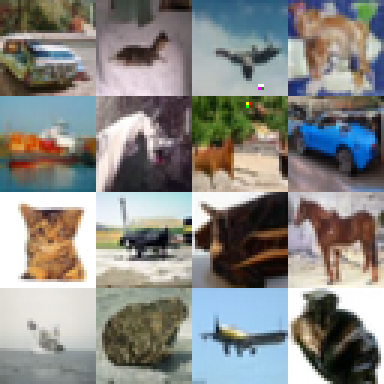}
    \end{subfigure} &
    \begin{subfigure}{0.15\textwidth}
        \includegraphics[width=\textwidth]{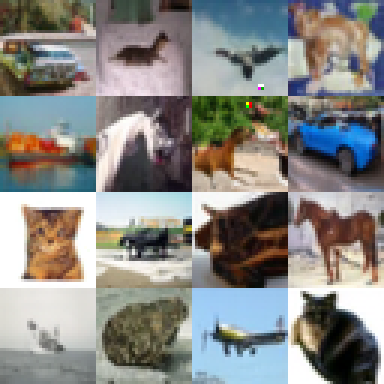}
    \end{subfigure} &
        \begin{subfigure}{0.15\textwidth}
        \includegraphics[width=\textwidth]{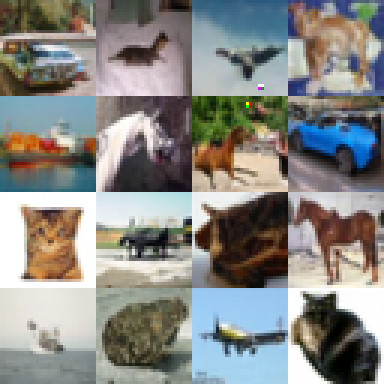}
    \end{subfigure} &
        \begin{subfigure}{0.15\textwidth}
        \includegraphics[width=\textwidth]{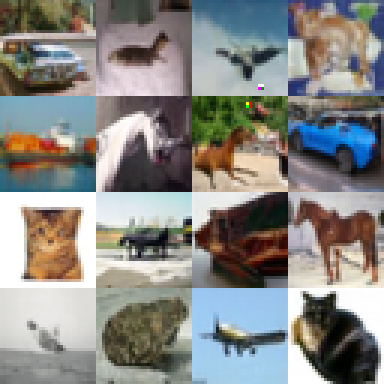}
    \end{subfigure} &
        \begin{subfigure}{0.15\textwidth}
        \includegraphics[width=\textwidth]{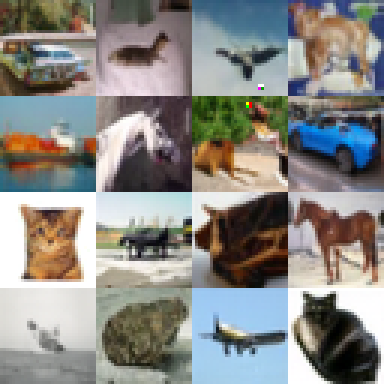}
    \end{subfigure} \\
    \raisebox{1.8\height}{\makecell{OFA Diffusion Compression \\ + Fine-tune 20K \\ Avg. FID = 5.22}} &
    \begin{subfigure}{0.15\textwidth}
        \includegraphics[width=\textwidth]{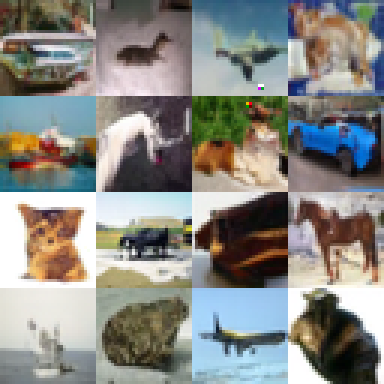}
    \end{subfigure} &
    \begin{subfigure}{0.15\textwidth}
        \includegraphics[width=\textwidth]{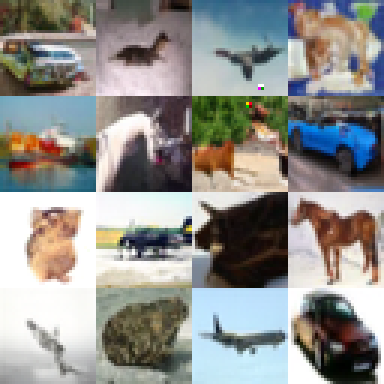}
    \end{subfigure} &
    \begin{subfigure}{0.15\textwidth}
        \includegraphics[width=\textwidth]{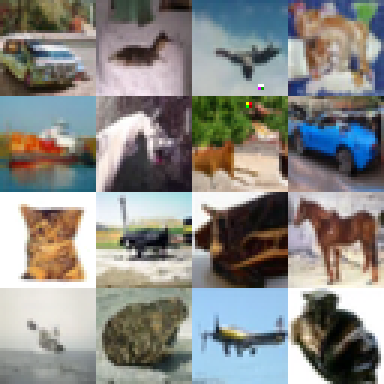}
    \end{subfigure} &
    \begin{subfigure}{0.15\textwidth}
        \includegraphics[width=\textwidth]{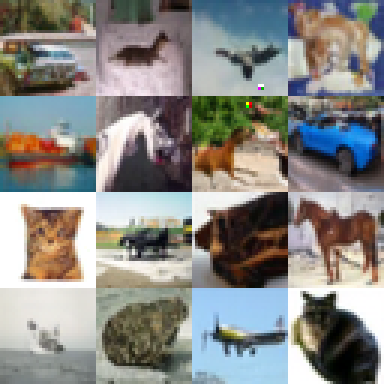}
    \end{subfigure} &
        \begin{subfigure}{0.15\textwidth}
        \includegraphics[width=\textwidth]{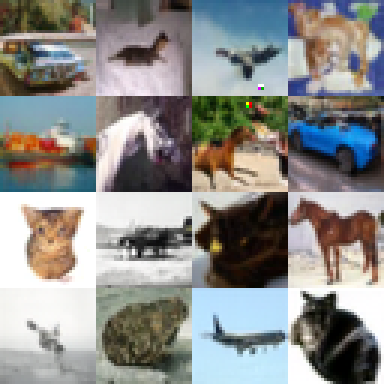}
    \end{subfigure} &
        \begin{subfigure}{0.15\textwidth}
        \includegraphics[width=\textwidth]{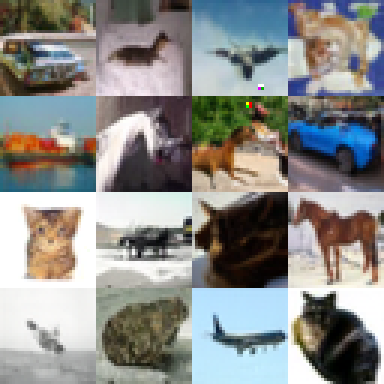}
    \end{subfigure} &
        \begin{subfigure}{0.15\textwidth}
        \includegraphics[width=\textwidth]{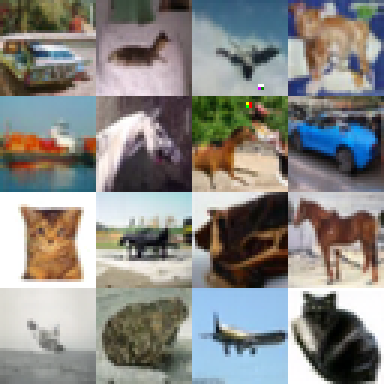}
    \end{subfigure} \\
    \end{tabular}
    }
    \caption{Samples from subnetworks trained by different approaches with the same random seed on cifar10 $32\times32$ (U-ViT unconditional sampling, Transformer backbone).}
\label{fig: vis_cifar10_uvit}
\end{figure*}

\begin{figure*}[!htb]
    \centering
    \setlength{\tabcolsep}{1pt}
    \resizebox{1.0\linewidth}{!}{
    \begin{tabular}{c c c c c c c c}
     & $\cP=0.25\times$ & $\cP=0.375\times$ & $\cP=0.5\times$ & $\cP=0.625\times$ & $\cP=0.75\times$ & $\cP=0.875\times$ & $\cP=1.0\times$\\
    \raisebox{2.0\height}{\makecell{Separate Compression \\ Avg. FID = 2.82}} &
    \begin{subfigure}{0.15\textwidth}
        \includegraphics[width=\textwidth]{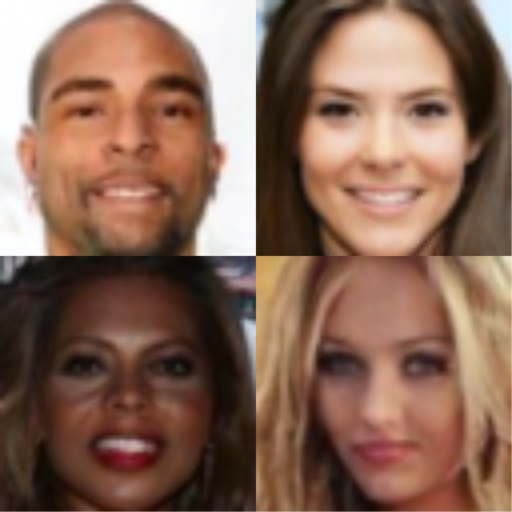}
    \end{subfigure} &
    \begin{subfigure}{0.15\textwidth}
        \includegraphics[width=\textwidth]{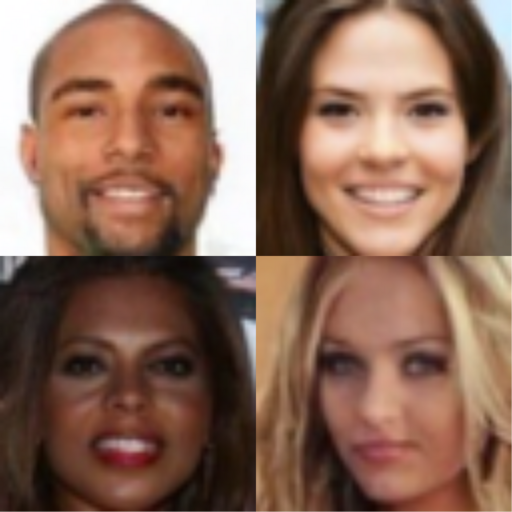}
    \end{subfigure} &
    \begin{subfigure}{0.15\textwidth}
        \includegraphics[width=\textwidth]{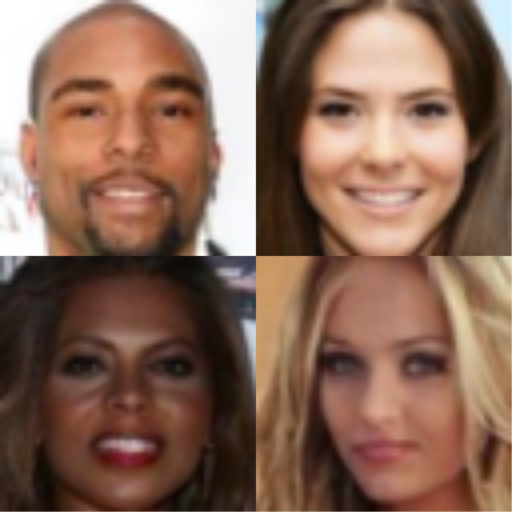}
    \end{subfigure} &
    \begin{subfigure}{0.15\textwidth}
        \includegraphics[width=\textwidth]{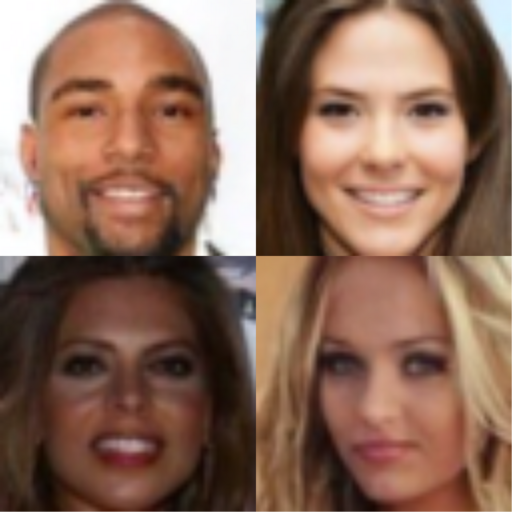}
    \end{subfigure} &
        \begin{subfigure}{0.15\textwidth}
        \includegraphics[width=\textwidth]{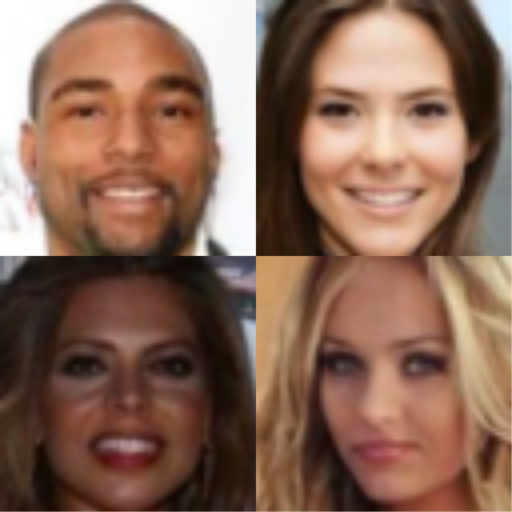}
    \end{subfigure} &
        \begin{subfigure}{0.15\textwidth}
        \includegraphics[width=\textwidth]{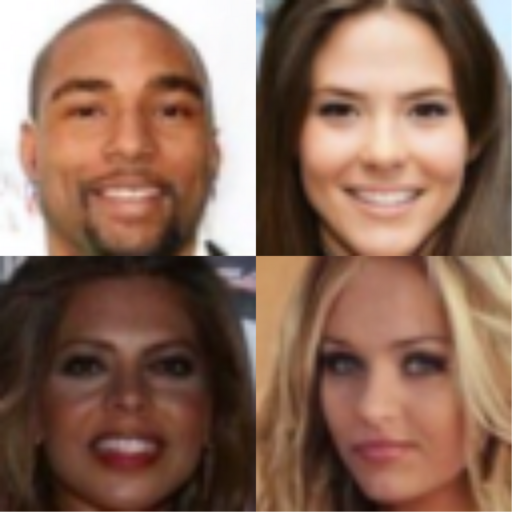}
    \end{subfigure} &
        \begin{subfigure}{0.15\textwidth}
        \includegraphics[width=\textwidth]{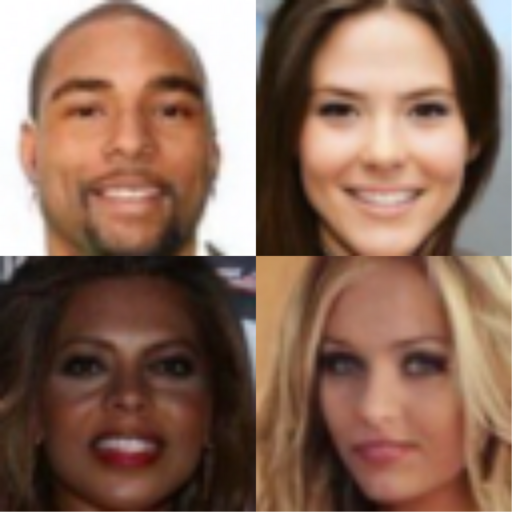}
    \end{subfigure} \\
    \raisebox{2.0\height}{\makecell{OFA Diffusion Compression \\ Avg. FID = 3.29}} &
    \begin{subfigure}{0.15\textwidth}
        \includegraphics[width=\textwidth]{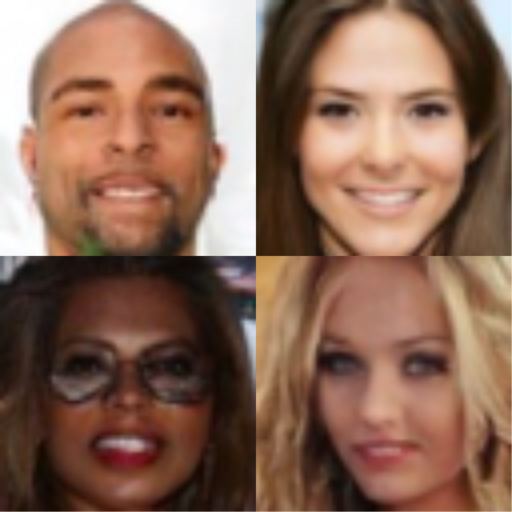}
    \end{subfigure} &
    \begin{subfigure}{0.15\textwidth}
        \includegraphics[width=\textwidth]{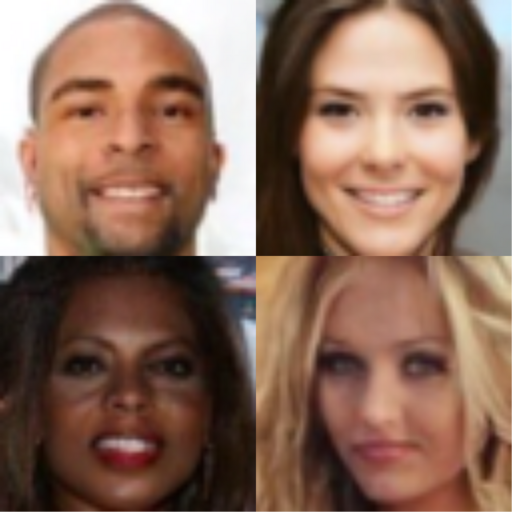}
    \end{subfigure} &
    \begin{subfigure}{0.15\textwidth}
        \includegraphics[width=\textwidth]{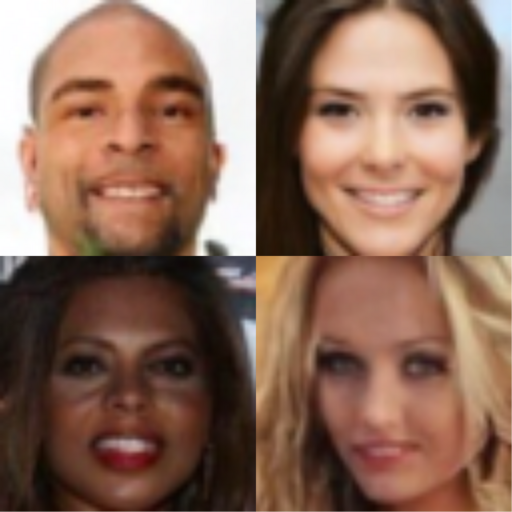}
    \end{subfigure} &
    \begin{subfigure}{0.15\textwidth}
        \includegraphics[width=\textwidth]{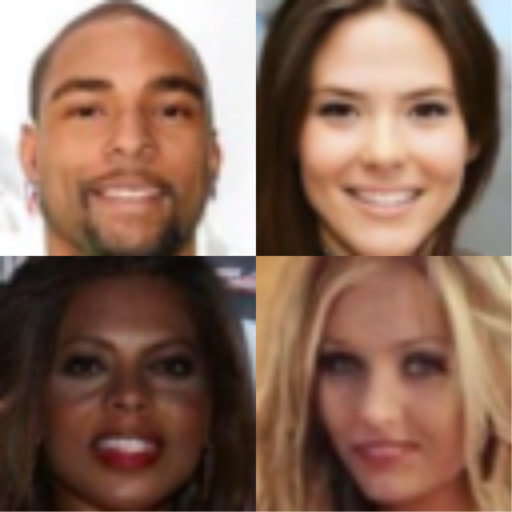}
    \end{subfigure} &
        \begin{subfigure}{0.15\textwidth}
        \includegraphics[width=\textwidth]{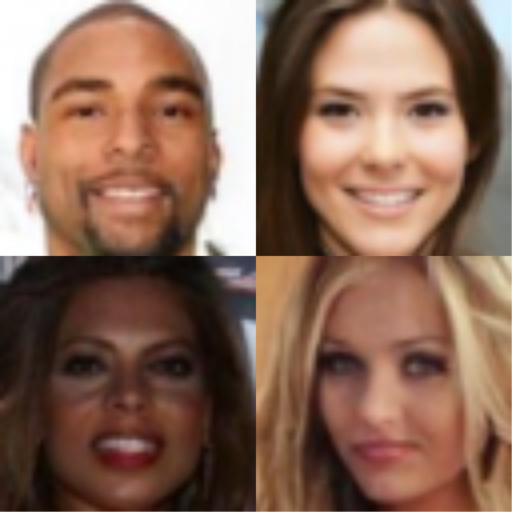}
    \end{subfigure} &
        \begin{subfigure}{0.15\textwidth}
        \includegraphics[width=\textwidth]{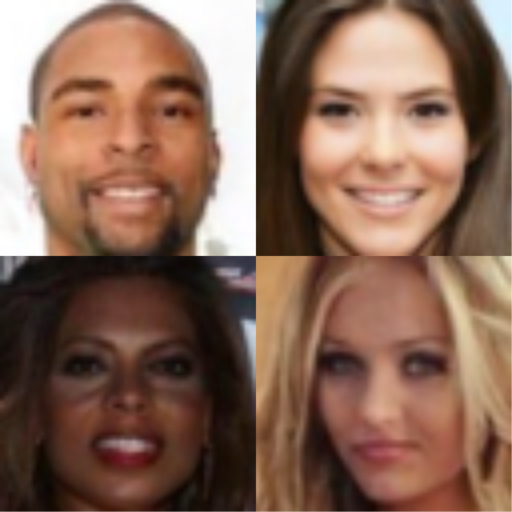}
    \end{subfigure} &
        \begin{subfigure}{0.15\textwidth}
        \includegraphics[width=\textwidth]{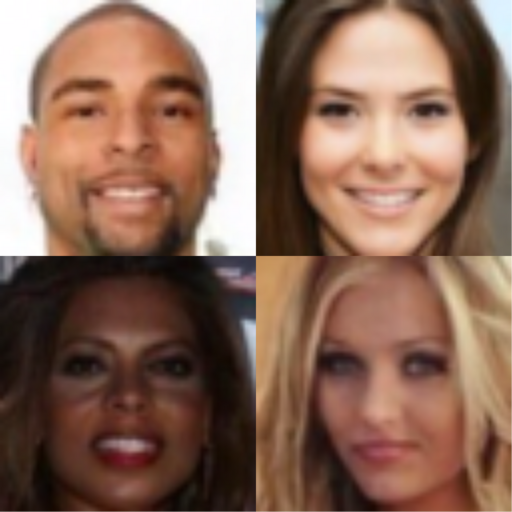}
    \end{subfigure} \\
    \raisebox{1.8 \height}{\makecell{OFA Diffusion Compression \\ +Fine-tune 10K \\ Avg. FID = 2.98}} &
    \begin{subfigure}{0.15\textwidth}
        \includegraphics[width=\textwidth]{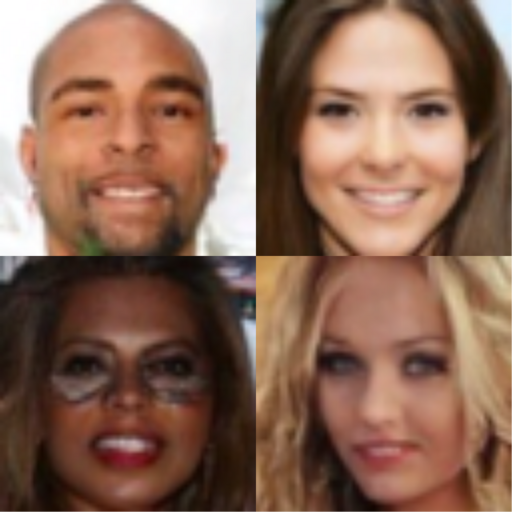}
    \end{subfigure} &
    \begin{subfigure}{0.15\textwidth}
        \includegraphics[width=\textwidth]{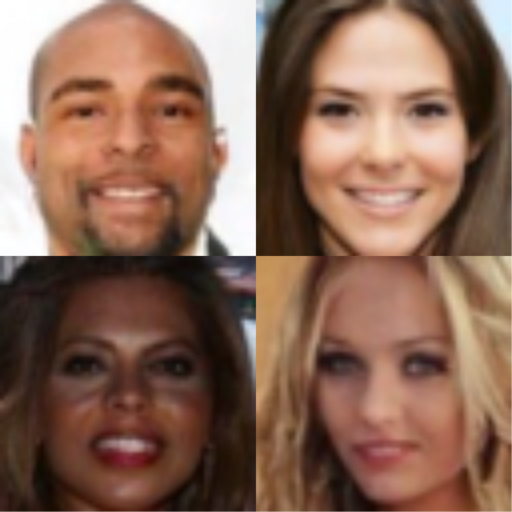}
    \end{subfigure} &
    \begin{subfigure}{0.15\textwidth}
        \includegraphics[width=\textwidth]{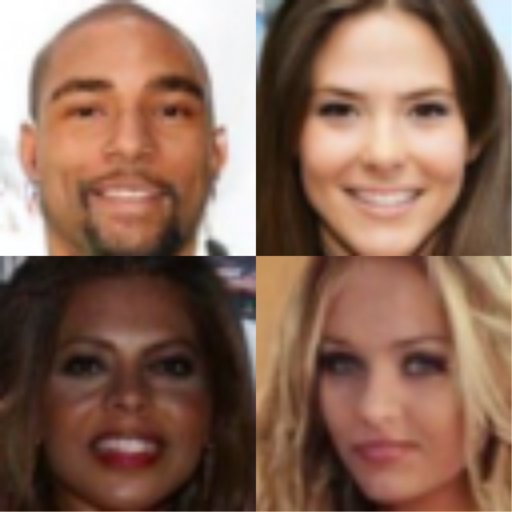}
    \end{subfigure} &
    \begin{subfigure}{0.15\textwidth}
        \includegraphics[width=\textwidth]{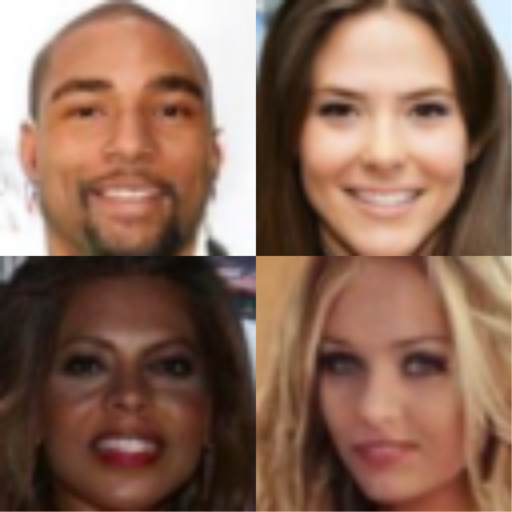}
    \end{subfigure} &
        \begin{subfigure}{0.15\textwidth}
        \includegraphics[width=\textwidth]{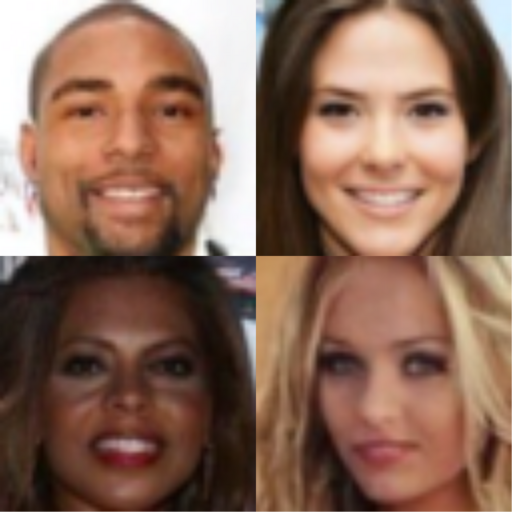}
    \end{subfigure} &
        \begin{subfigure}{0.15\textwidth}
        \includegraphics[width=\textwidth]{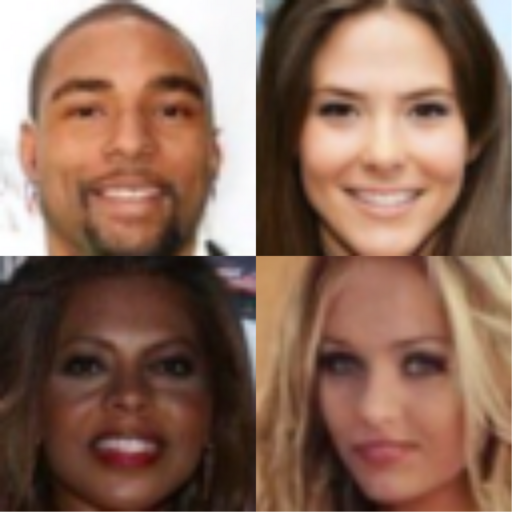}
    \end{subfigure} &
        \begin{subfigure}{0.15\textwidth}
        \includegraphics[width=\textwidth]{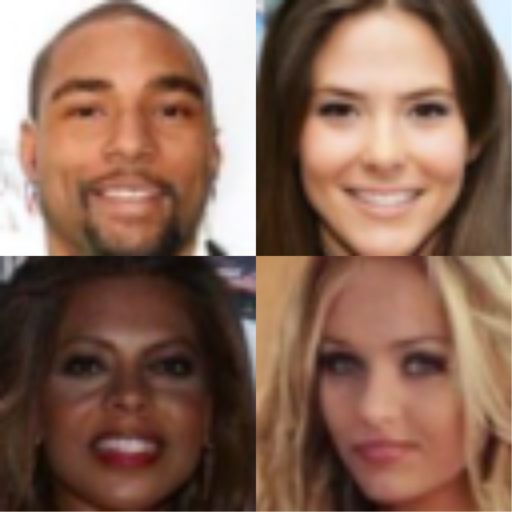}
    \end{subfigure} \\
    \raisebox{1.8\height}{\makecell{OFA Diffusion Compression \\ + Fine-tune 20K \\ Avg. FID = 2.78}} &
    \begin{subfigure}{0.15\textwidth}
        \includegraphics[width=\textwidth]{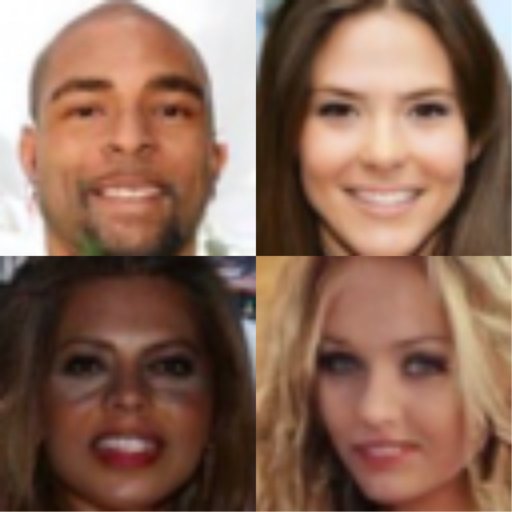}
    \end{subfigure} &
    \begin{subfigure}{0.15\textwidth}
        \includegraphics[width=\textwidth]{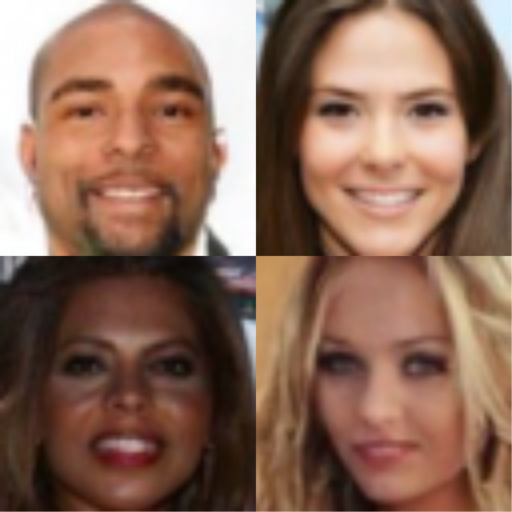}
    \end{subfigure} &
    \begin{subfigure}{0.15\textwidth}
        \includegraphics[width=\textwidth]{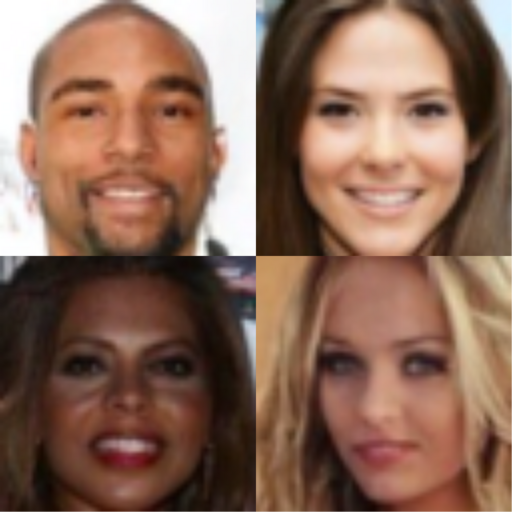}
    \end{subfigure} &
    \begin{subfigure}{0.15\textwidth}
        \includegraphics[width=\textwidth]{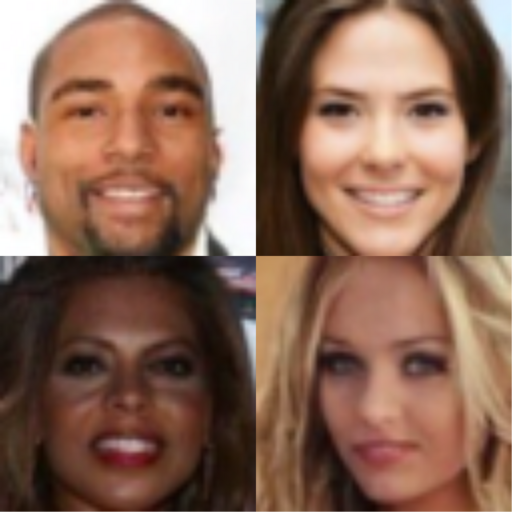}
    \end{subfigure} &
        \begin{subfigure}{0.15\textwidth}
        \includegraphics[width=\textwidth]{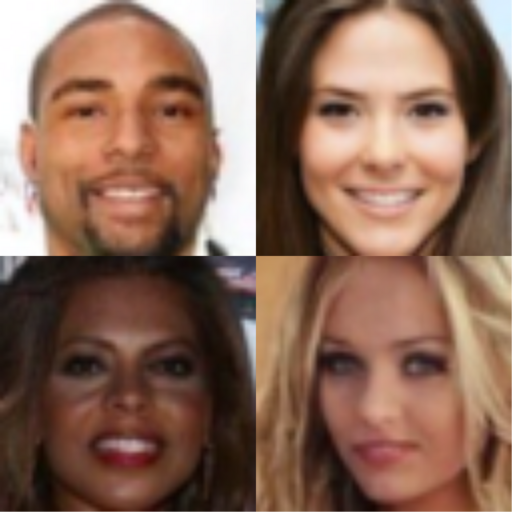}
    \end{subfigure} &
        \begin{subfigure}{0.15\textwidth}
        \includegraphics[width=\textwidth]{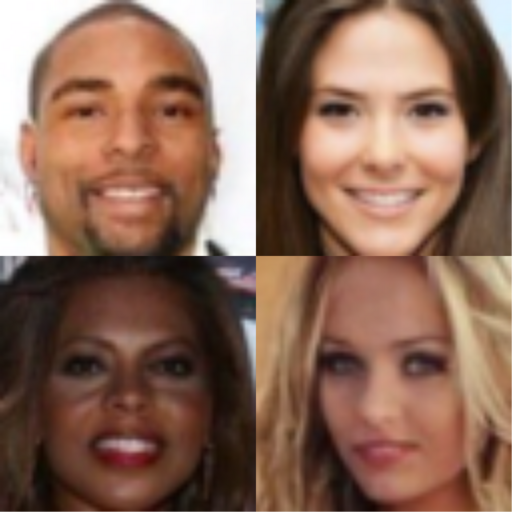}
    \end{subfigure} &
        \begin{subfigure}{0.15\textwidth}
        \includegraphics[width=\textwidth]{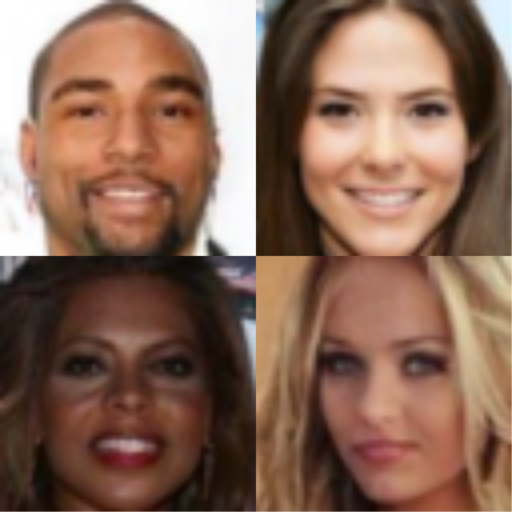}
    \end{subfigure} \\
    \end{tabular}
    }
    \caption{Samples from subnetworks trained by different approaches with the same random seed on CelebA $64\times64$ (U-ViT unconditional sampling, Transformer backbone).}
\label{fig: vis_celeba}
\end{figure*}

\begin{figure*}[!htb]
    \centering
    \setlength{\tabcolsep}{1pt}
    \resizebox{1.0\linewidth}{!}{
    \begin{tabular}{c c c c c c c c}
     & $\cP=0.25\times$ & $\cP=0.375\times$ & $\cP=0.5\times$ & $\cP=0.625\times$ & $\cP=0.75\times$ & $\cP=0.875\times$ & $\cP=1.0\times$\\
    \raisebox{2.0\height}{\makecell{Separate Compression \\ Avg. FID = 10.10}} &
    \begin{subfigure}{0.15\textwidth}
        \includegraphics[width=\textwidth]{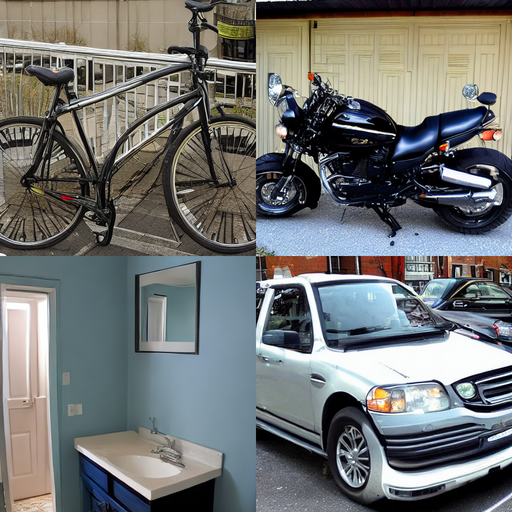}
    \end{subfigure} &
    \begin{subfigure}{0.15\textwidth}
        \includegraphics[width=\textwidth]{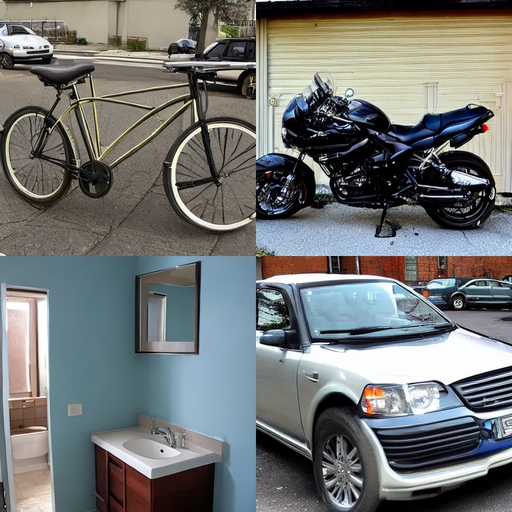}
    \end{subfigure} &
    \begin{subfigure}{0.15\textwidth}
        \includegraphics[width=\textwidth]{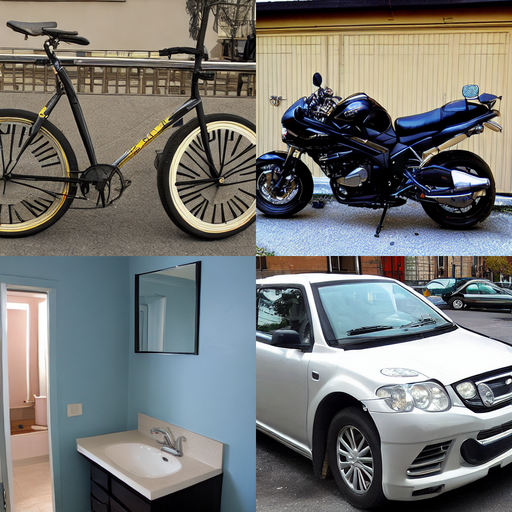}
    \end{subfigure} &
    \begin{subfigure}{0.15\textwidth}
        \includegraphics[width=\textwidth]{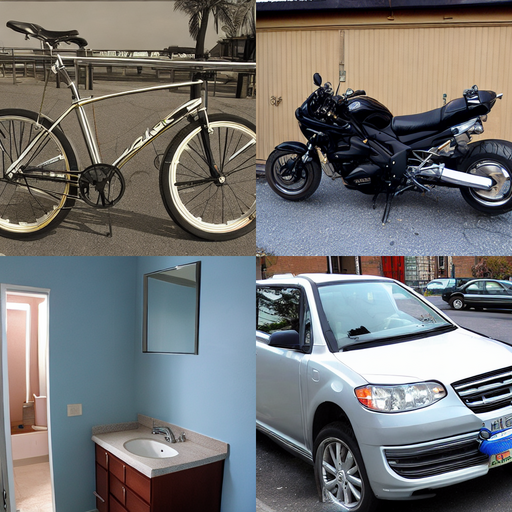}
    \end{subfigure} &
        \begin{subfigure}{0.15\textwidth}
        \includegraphics[width=\textwidth]{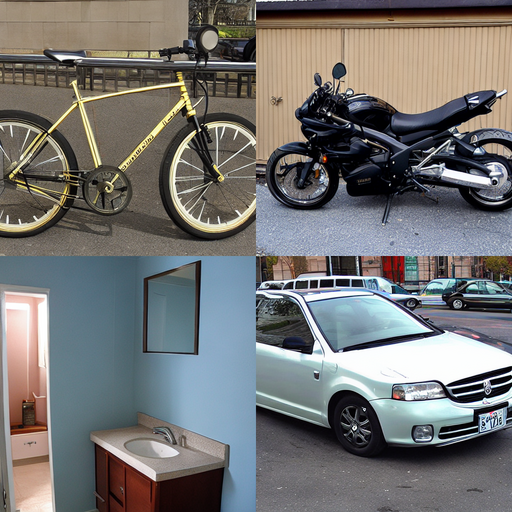}
    \end{subfigure} &
        \begin{subfigure}{0.15\textwidth}
        \includegraphics[width=\textwidth]{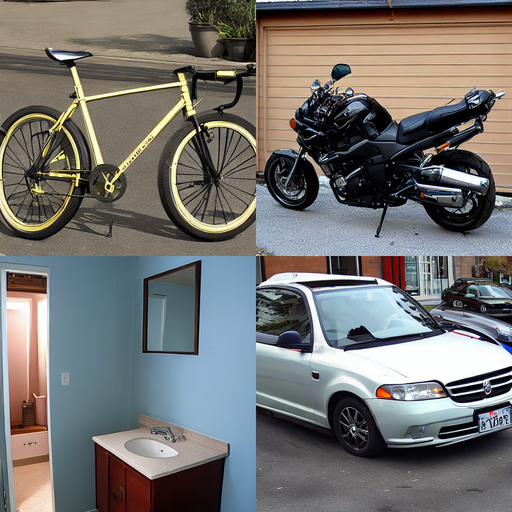}
    \end{subfigure} &
        \begin{subfigure}{0.15\textwidth}
        \includegraphics[width=\textwidth]{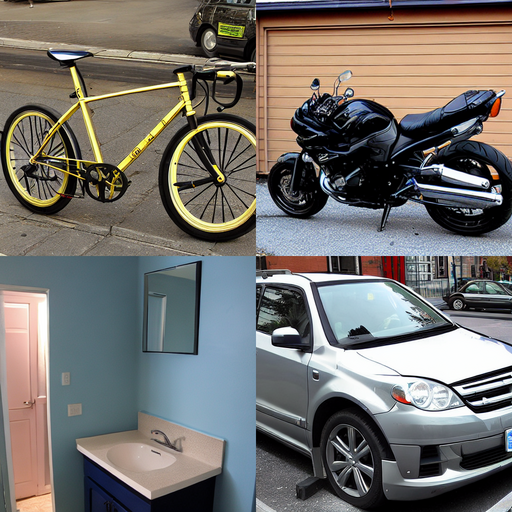}
    \end{subfigure} \\
    \raisebox{2.0\height}{\makecell{OFA Diffusion Compression \\ Avg. FID = 8.85}} &
    \begin{subfigure}{0.15\textwidth}
        \includegraphics[width=\textwidth]{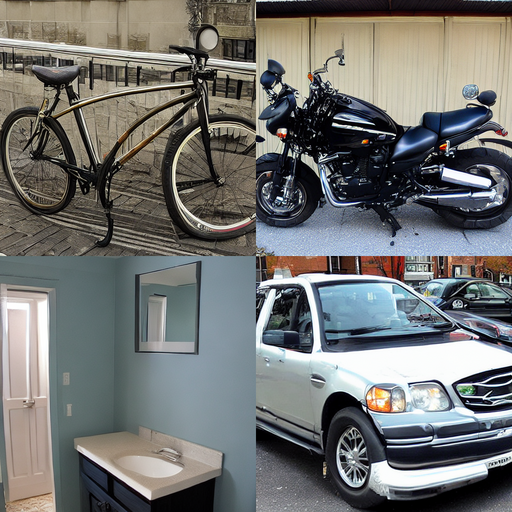}
    \end{subfigure} &
    \begin{subfigure}{0.15\textwidth}
        \includegraphics[width=\textwidth]{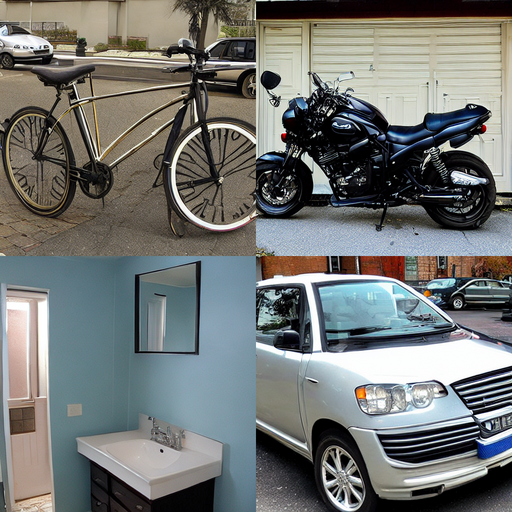}
    \end{subfigure} &
    \begin{subfigure}{0.15\textwidth}
        \includegraphics[width=\textwidth]{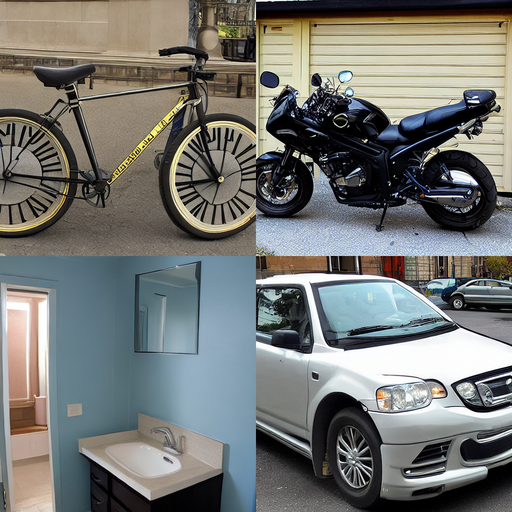}
    \end{subfigure} &
    \begin{subfigure}{0.15\textwidth}
        \includegraphics[width=\textwidth]{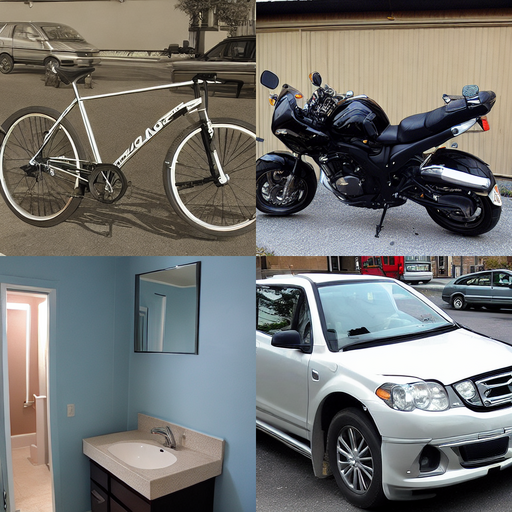}
    \end{subfigure} &
        \begin{subfigure}{0.15\textwidth}
        \includegraphics[width=\textwidth]{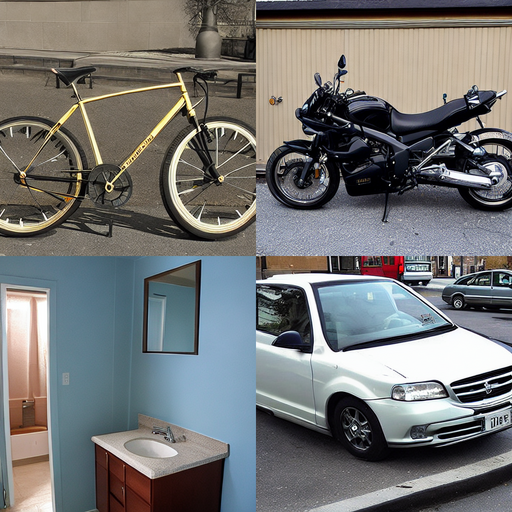}
    \end{subfigure} &
        \begin{subfigure}{0.15\textwidth}
        \includegraphics[width=\textwidth]{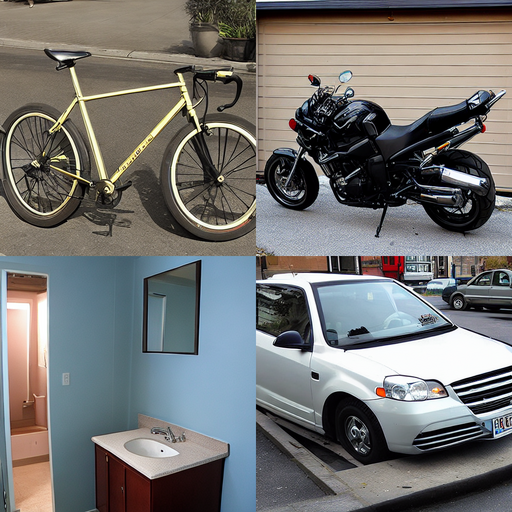}
    \end{subfigure} &
        \begin{subfigure}{0.15\textwidth}
        \includegraphics[width=\textwidth]{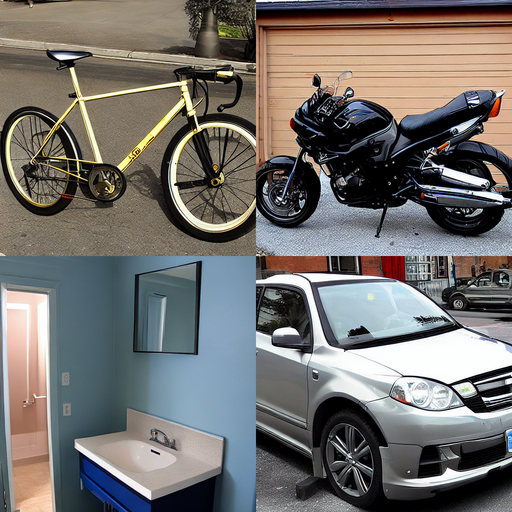}
    \end{subfigure} \\
    \raisebox{1.8 \height}{\makecell{OFA Diffusion Compression \\ +Fine-tune 10K \\ Avg. FID = 9.30}} &
    \begin{subfigure}{0.15\textwidth}
        \includegraphics[width=\textwidth]{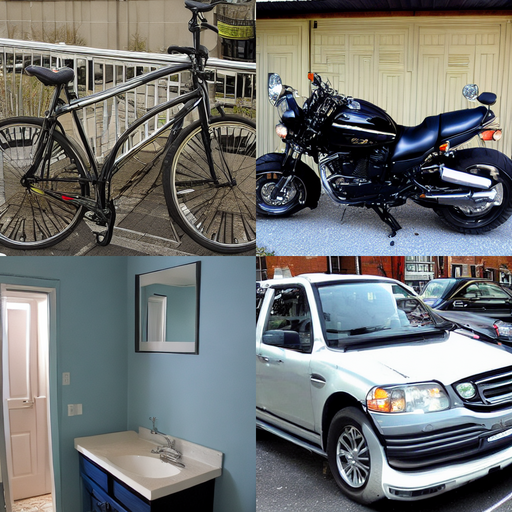}
    \end{subfigure} &
    \begin{subfigure}{0.15\textwidth}
        \includegraphics[width=\textwidth]{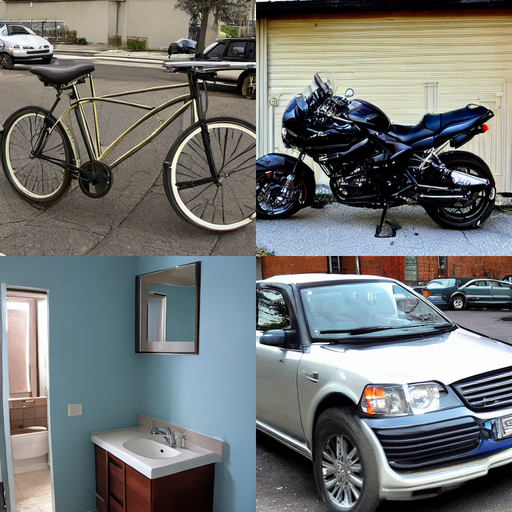}
    \end{subfigure} &
    \begin{subfigure}{0.15\textwidth}
        \includegraphics[width=\textwidth]{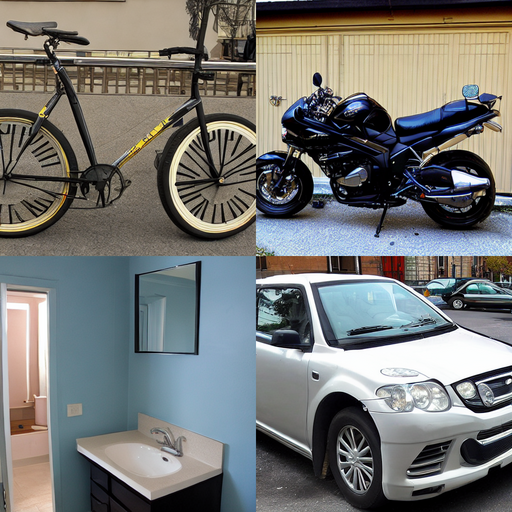}
    \end{subfigure} &
    \begin{subfigure}{0.15\textwidth}
        \includegraphics[width=\textwidth]{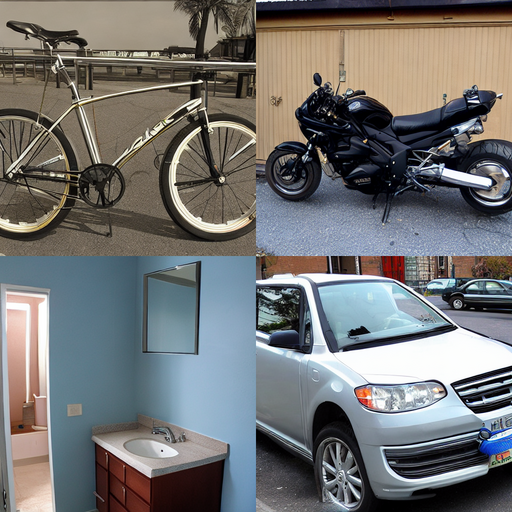}
    \end{subfigure} &
        \begin{subfigure}{0.15\textwidth}
        \includegraphics[width=\textwidth]{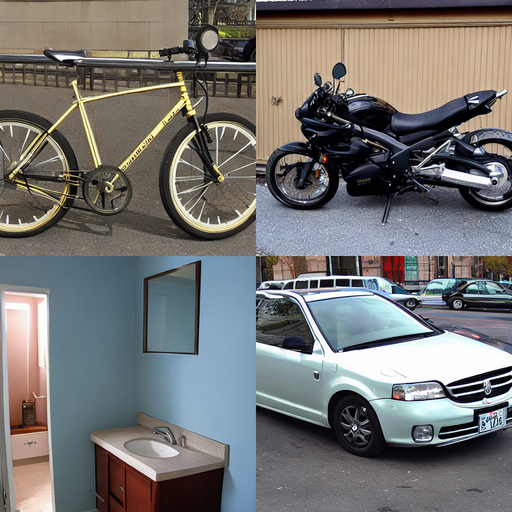}
    \end{subfigure} &
        \begin{subfigure}{0.15\textwidth}
        \includegraphics[width=\textwidth]{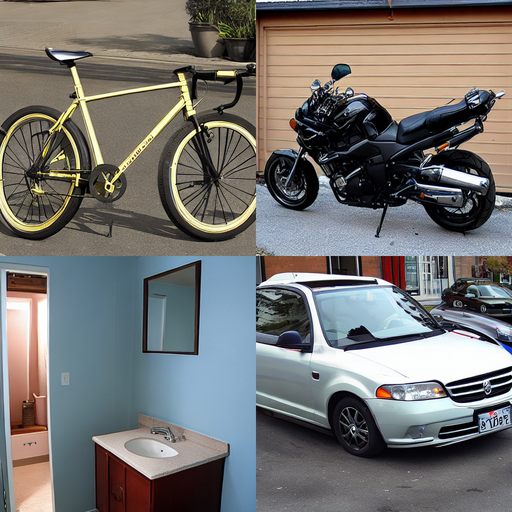}
    \end{subfigure} &
        \begin{subfigure}{0.15\textwidth}
        \includegraphics[width=\textwidth]{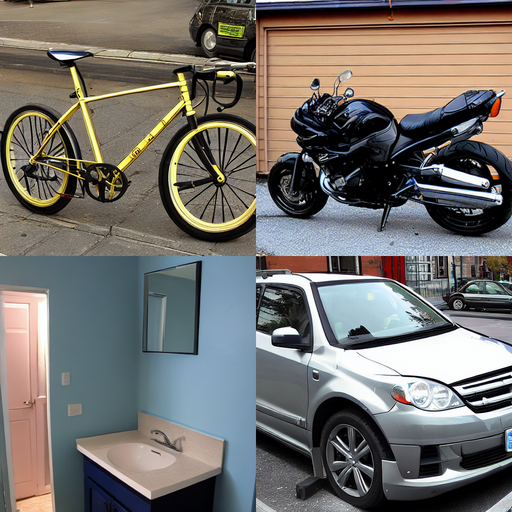}
    \end{subfigure} \\
    \raisebox{1.8\height}{\makecell{OFA Diffusion Compression \\ + Fine-tune 20K \\ Avg. FID = 9.67}} &
    \begin{subfigure}{0.15\textwidth}
        \includegraphics[width=\textwidth]{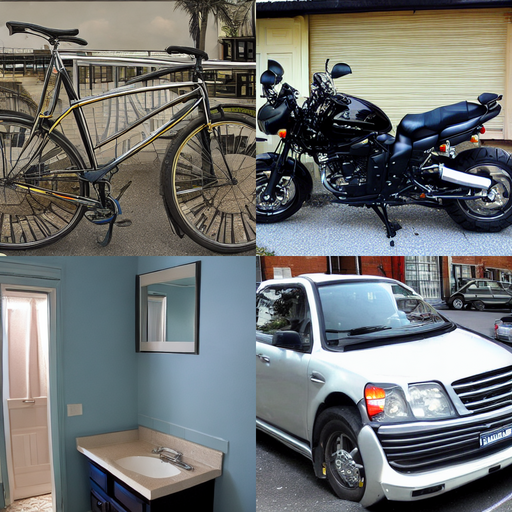}
    \end{subfigure} &
    \begin{subfigure}{0.15\textwidth}
        \includegraphics[width=\textwidth]{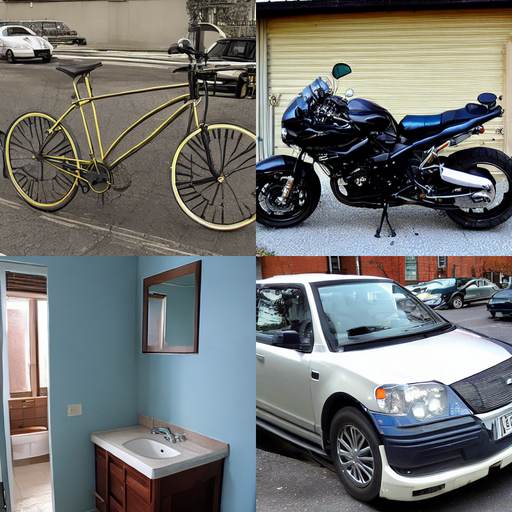}
    \end{subfigure} &
    \begin{subfigure}{0.15\textwidth}
        \includegraphics[width=\textwidth]{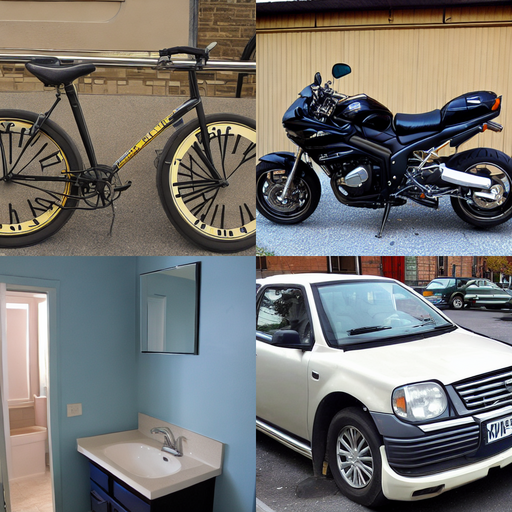}
    \end{subfigure} &
    \begin{subfigure}{0.15\textwidth}
        \includegraphics[width=\textwidth]{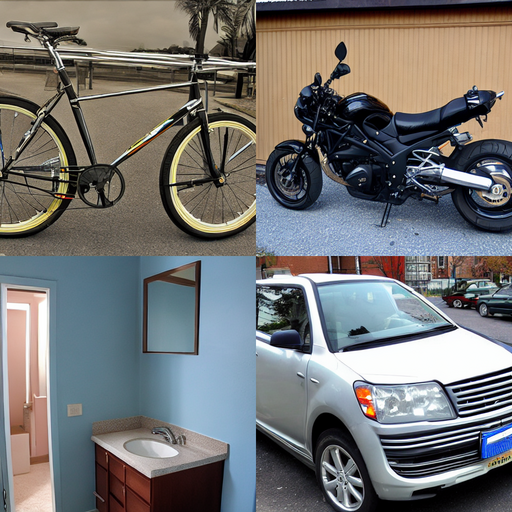}
    \end{subfigure} &
        \begin{subfigure}{0.15\textwidth}
        \includegraphics[width=\textwidth]{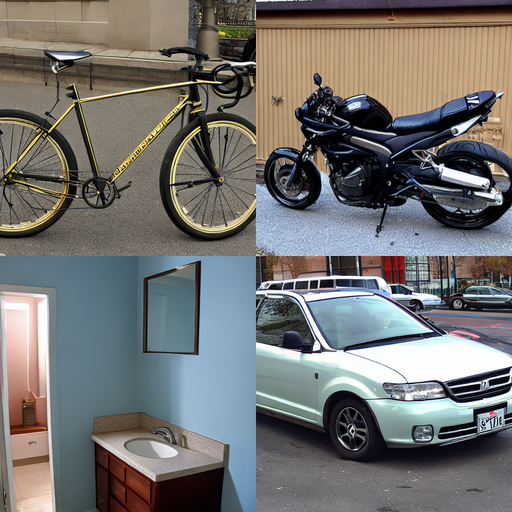}
    \end{subfigure} &
        \begin{subfigure}{0.15\textwidth}
        \includegraphics[width=\textwidth]{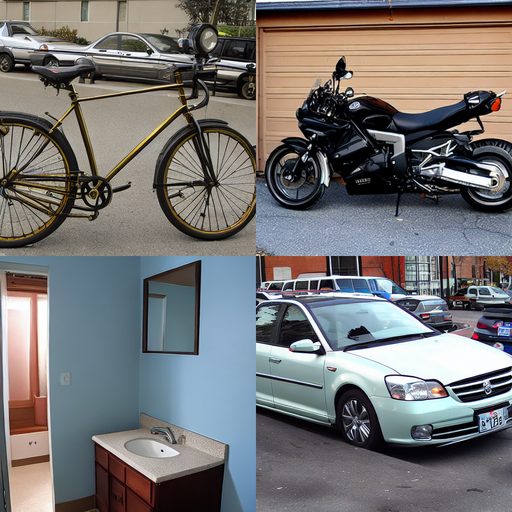}
    \end{subfigure} &
        \begin{subfigure}{0.15\textwidth}
        \includegraphics[width=\textwidth]{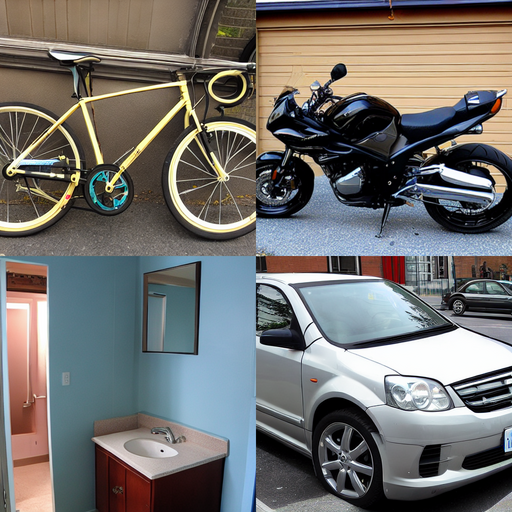}
    \end{subfigure} \\
    \end{tabular}
    }
    \caption{Samples from subnetworks trained by different approaches with the same random seed on MS-COCO $512\times512$ (Stable Diffusion text-to-image generation).}
\label{fig: vis_mscoco}
\end{figure*}

In this section, we provide random samples from subnetworks trained by different approaches under various $\cP_{i}$, as shown in Figures~\ref{fig: vis_cifar10}, \ref{fig: vis_ffhq}, \ref{fig: vis_afhqv2}, \ref{fig: vis_imagenet}, \ref{fig: vis_cifar10_uvit}, \ref{fig: vis_celeba}, and \ref{fig: vis_mscoco}. 
We observe that the subnetworks extracted from our OFA network can generate high-quality samples with clear semantics.


\clearpage

\end{document}